\documentclass[pdflatex,sn-basic,Numbered,iicol]{sn-jnl}

\usepackage{graphicx}%
\usepackage{multirow}%
\usepackage{amsmath,amssymb,amsfonts}%
\usepackage[title]{appendix}%
\usepackage{xcolor}%
\usepackage{textcomp}%
\usepackage{manyfoot}%
\usepackage{booktabs}%
\usepackage{tabularx,makecell,array}%
\usepackage{pifont}%
\usepackage{placeins}%
\newcommand{\fitwidth}[1]{\resizebox{\ifdim\width>\textwidth\textwidth\else\width\fi}{!}{#1}}
\usepackage{needspace}
\usepackage[most]{tcolorbox}%
\newtcolorbox{promptbox}{breakable, enhanced, colback=yellow!6, colframe=black!25, boxrule=0.4pt, arc=1.5pt, left=4pt, right=4pt, top=3pt, bottom=3pt, fontupper=\scriptsize}

\newcommand{\req}{\ding{108}}            
\newcommand{\pinf}{\ding{109}}           
\newcommand{\nr}{--}                     
\newcolumntype{L}[1]{>{\raggedright\arraybackslash}p{#1}}
\newcolumntype{Y}{>{\raggedright\arraybackslash}X}

\usepackage{fancyvrb}
\newcommand{\promptfont}{\fontsize{8.5}{10}\selectfont}

\makeatletter
\newenvironment{appendixtable}{\par\addvspace{\intextsep}\noindent\begin{minipage}{\textwidth}\def\@captype{table}}{\end{minipage}\par\addvspace{\intextsep}}
\newenvironment{appendixfigure}{\par\addvspace{\intextsep}\noindent\begin{minipage}{\textwidth}\def\@captype{figure}}{\end{minipage}\par\addvspace{\intextsep}}
\makeatother
\gdef\orcidlogo{\textcolor[HTML]{A6CE39}{\textsf{\textbf{\scriptsize iD}}}}
\let\originalprintkeywords\printkeywords
\renewcommand{\printkeywords}{\originalprintkeywords\par\medskip
  {\keywordfont\parfillskip=0pt plus 1fil\noindent\textbf{Acknowledgements.} We thank the audiences at COMPTEXT 2026 and the 2026 annual meeting of the Methods Division of the German Communication Association (DGPuK) for helpful comments on earlier versions of this manuscript.\par}}
\begin{document}

\title[What Does the Model See?]{What Does the Model See? Representation Validity in Multimodal LLM-Based Video Analysis}

\author*[1]{\fnm{Christian} \sur{Pipal}\,\orcid{https://orcid.org/0000-0002-5395-2035}}\email{c.pipal@ikmz.uzh.ch}
\author[1]{\fnm{Eva-Maria} \sur{Vogel}\,\orcid{https://orcid.org/0009-0009-7593-1081}}
\author[1]{\fnm{Morgan} \sur{Wack}\,\orcid{https://orcid.org/0000-0002-7769-5993}}

\affil[1]{\orgdiv{Department of Communication and Media Research}, \orgname{University of Zurich}, \orgaddress{\city{Zurich}, \country{Switzerland}}}

\abstract{Multimodal large language models (MLLMs) allow social scientists to analyze video at scale, but researchers must decide which parts of a video to provide as input. We use the term representation validity to examine how text, audio, video, and combined inputs affect measurement and subsequent analysis. We compare input representations across six open-weight MLLMs and three human-annotated video datasets. No representation consistently performs best. On seven news-related constructs, native-video performance varies little across models (macro-F1 = .70--.77), while extracted-text performance ranges from .52 to .80. Three models perform better with text and three with video; combining inputs rarely improves performance. Higher classification accuracy also does not reliably produce estimates closer to those based on human codes. The highest-F1 representation best reproduces human-coded regression coefficients in 17 of 42 comparisons. Even when each model uses its highest-F1 deployable representation, 45\% of sample-prevalence estimates differ from the human-coded benchmark by more than ten percentage points. Input representation therefore requires validation for both the construct being measured and the quantity estimated from the resulting labels.}

\keywords{large language models, multimodal annotation, measurement validity, video, computational social science, content analysis}

\maketitle

\section{Introduction}\label{sec:intro}

Before a large language model can code a video, researchers must decide how to present it. They can supply a speech transcript, add optical character recognition (OCR) of on-screen text, sample frames and attach audio, use the model's video pathway, upload the original file, or ask the model to describe the clip and then code that description. These choices determine which information the model can use to measure the construct of interest.

So far, input representation has received less attention than model and prompt selection. LLM annotation is now common in computational social science. Zero-shot models can match crowd workers and, on some tasks, trained experts \citep{Gilardietal2023chatgpt,Tornberg2023chatgpt,HeseltineClemm2024large}. They work across languages and constructs \citep{Rathjeetal2024gpt}, and open-weight models allow researchers to run versioned annotation pipelines locally \citep{Alizadehetal2025opensource}. Their labels support claims about political communication, engagement, and elite behavior \citep{Marksetal2025ingroup,Solovevetal2026tiktok,TornbergChueri2025incivility}. Validation studies to date have focused on which models, prompts, and decoding settings produce agreement with human coders \citep{Ziemsetal2024can,Abrahametal2025prompt}. For text annotation, the input often closely resembles the document researchers want to classify. Video requires additional decisions about which channels to retain and how to encode them.

A thirty-second TikTok news video may convey meaning through speech, on-screen headlines, footage, editing, music, or delivery. A transcript retains the words but discards the picture and most properties of delivery. OCR recovers written content but not music or visual context. A sparse frame sample may capture the setting while missing a brief end card. A model's native video pathway can preserve temporal structure, although frame sampling and audio alignment vary by architecture. The value of each input depends on whether it preserves information relevant to the construct.

Let $X$ denote the original video, $R$ the procedure used to represent it, and $X_R=R(X)$ the model input. A model $f_\theta$ produces
\begin{equation}
\hat{Y}=f_\theta\big(R(X)\big).
\label{eq:pipeline}
\end{equation}
Varying $R$ while holding the model and prompt constant isolates the effect of representation on measurement. Researchers using the same model and codebook can obtain different variables from the same videos if they supply different inputs.

We use \emph{representation validity} to describe this part of the measurement process within established approaches to construct validity. We examine whether the representation retains the information a construct requires, whether the model uses that information, whether it combines complementary signals, and whether the resulting variable preserves the quantity researchers want to estimate.

We examine these questions using six open-weight audio-visual LLMs and three datasets with human reference labels. We use 603 German-language TikTok news videos with codes for news values, platform features, and audiovisual properties \citep{Mayeretal2025newstiktok}; 400 English-language videos with codes for hate and for whether they target Black or Jewish people \citep{Das2023hatemm}; and 650 English sitcom clips with sarcasm labels \citep{Castro2019mustard}. We give each model the same codebook instructions while varying the input representation. Across the datasets, we compare nine input routes: metadata only, an automatic transcript, extracted speech and on-screen text, a human transcript where available, sampled frames with audio, native video, full video, combined text and video, and a model-written description. We remove audio to test whether models use sound, and we compare errors across routes to test whether models combine complementary text and video signals. We then re-estimate engagement regressions and sample prevalences to assess how coding accuracy relates to downstream estimates. Finally, we repeat runs to check computational stability.

On the seven verbally defined news constructs, native-video macro-F1 ranges from .70 to .77 across models, compared with .52 to .80 for extracted text. Three models perform better with extracted text and three with native video. The wider comparison also shows that representation effects vary by model and construct. Comparing constructs on the same videos helps separate these effects from differences between datasets: audiovisual input improves coding of visual setup more consistently than coding of verbal news constructs, and its advantage differs between the two HateMM target-group variables.

The audio ablations show that models use sound differently across tasks. Removing audio can substantially reduce performance on hate or sarcasm while barely affecting another task. Both Gemma models can transcribe English speech with an appropriate prompt, yet rarely identify music. Text and audiovisual routes also make complementary errors, but providing both in one call usually yields little improvement over the better individual route.

Higher macro-F1 does not reliably identify the representation that best recovers a downstream estimate. In the news application, the highest-F1 route produces the smallest regression-coefficient error in 17 of 42 model--construct comparisons. Across all three datasets, 45\% of sample-prevalence estimates remain more than ten percentage points from the human-coded reference even after selecting the highest-F1 deployable route. These results support validating model--representation combinations for the intended construct and for the estimates calculated from their labels.

\section{Related work}\label{sec:related}

\subsection{LLM annotation in computational social science}

Automated content analysis is a measurement problem: an instrument should agree with a trusted reference and support valid substantive inference \citep{GrimmerStewart2013text,Songetal2020validations,Badenetal2022gaps}. Generative LLMs have made this work less costly. Zero-shot annotation with ChatGPT outperformed crowd workers on several political-text tasks \citep{Gilardietal2023chatgpt}, matched or exceeded experts on political tweets \citep{Tornberg2023chatgpt}, and approached expert coders across political-science annotation tasks \citep{HeseltineClemm2024large}. Similar results hold for sentiment, discrete emotions, and offensiveness across twelve languages \citep{Rathjeetal2024gpt}, although performance varies across computational social science tasks and constructs \citep{Ziemsetal2024can}.

Subsequent work has examined how to produce and validate these annotations. Open-weight models permit local processing and reproduction from archived weights, but researchers must report model settings and fine-tuning choices that affect performance \citep{Alizadehetal2025opensource,Ollionetal2024dangers}. Small changes in prompt wording can also shift accuracy \citep{Abrahametal2025prompt}. Validation must extend beyond classification: even accurate labels can bias downstream regressions, and methods such as design-based supervised learning use human-coded subsamples to correct that bias \citep{TeBlunthuisetal2024misclassification,Egamietal2023surrogates}. Operational choices affect costs as well; annotating multiple items and variables per call can reduce costs with little loss of accuracy \citep{Pipaletal2026batching}.

Substantive studies increasingly use LLM-generated codes as measures. GPT-4o codes of ingroup solidarity and outgroup hostility in Facebook posts show how engagement shifted after the political crises of the 2024 US campaign \citep{Marksetal2025ingroup}; LLM measures of incivility and character attacks link hostile campaign rhetoric to electoral outcomes \citep{BanksStone2026hostile}; LLM-coded tweets of parliamentarians in 26 countries trace a rise in elite incivility \citep{TornbergChueri2025incivility}; and language-model estimates track outgroup animosity among politicians, journalists, and partisan users over a decade \citep{FulayRoy2026outgroup}. Similar designs measure framing in news articles \citep{Tohidietal2025framing} and mobilization in protest communication \citep{Eckardtetal2025protest}. Increasingly, the material is audiovisual. A study of 25{,}292 TikTok videos by German politicians uses GPT-4-turbo to code automatic video transcripts for outgroup animosity, identity language, and relatability \citep{Solovevetal2026tiktok}. Multimodal models now also code who appears in candidates' Instagram stories \citep{AchmannDenkleretal2026instagram}. These studies depend on the representation supplied to the model, although they rarely compare alternative representations.

Model and prompt selection thus have established roles in annotation validation. For multimodal data, researchers must also decide how to represent the material before the model processes it.

\subsection{From text annotation to multimodal social data}

Research that codes audiovisual political material has historically combined speech recognition for what is said, computer vision for faces, objects, and settings, and human coding or supervised classifiers on top \citep{JooSteinert2022image,Boussalisetal2021gender}. A validation of automated coding for campaign advertisements shows that such component pipelines reach human-level reliability for many but not all variables \citep{TarrHwangImai2023automated}. Which channel carries a construct matters. When human coders rate parliamentary speeches from transcripts and from video, the valence of a speech carries over to the transcript, whereas its emotional arousal does not \citep{Cochrane2022emotion}.

Multimodal LLMs can process several channels in one call, but their performance varies across settings. Open and proprietary multimodal models approach human reliability when rating emotional arousal in laboratory recordings but fall well short on real parliamentary debates \citep{Licht2026mllm}. Benchmarks for video LLMs show that models can answer many items without the visual channel and that accuracy does not necessarily reflect use of that channel \citep{Lee2026grounding}. Video LLMs also read on-screen text poorly \citep{Shietal2025mmevideoocr}, which matters for platforms on which much of a video's verbal content appears as overlaid text. Converting video into text first can sometimes improve performance. For example, describing clips and reasoning over the descriptions outperforms end-to-end video models on long-video question answering \citep{Zhangetal2024llovi}.

This work shows that constructs rely on different channels, models vary in their ability to extract them, and intermediate text representations can perform well. It leaves open how alternative representations compare when researchers hold the videos, social-science constructs, model, and prompt constant.

\subsection{Representation as a measurement decision}

Equation~\eqref{eq:pipeline} distinguishes the original object $X$, its representation $X_R$, and the model $f_\theta$. Holding the model, prompt, and decoding settings constant allows us to compare representations using four checks.

\emph{Representation sufficiency} concerns $R$. A representation is sufficient for a construct if it retains the information that the construct definition requires. A transcript cannot be sufficient for music detection, and sixteen sampled frames may miss a brief on-screen prompt. Researchers can assess sufficiency from each construct's codebook definition before running a model (Table~\ref{tab:requirements}).

\emph{Representation utilization} concerns $f_\theta$ given $X_R$. A model may receive information it does not use. We test this by removing a channel from an otherwise identical input and checking whether the model's codes change.

\emph{Representation integration} concerns $f_\theta$ given several representations at once. When two representations make different errors, a model that receives both could in principle do better than either. Whether it does is an empirical question.

\emph{Inferential validity} concerns estimates calculated from $\hat{Y}$. Automated labels can agree closely with human codes yet yield different regression coefficients, because the effect of a coding error depends on where it occurs in the data \citep{TeBlunthuisetal2024misclassification}.

We use these four checks to define our research questions. \textbf{RQ1} asks how much annotation validity changes when we present the same videos through different representations. \textbf{RQ2} asks whether representation effects depend on the information requirements of the target construct. \textbf{RQ3} asks whether multimodal models use the modalities we provide and whether they can integrate complementary signals. \textbf{RQ4} asks whether representation-induced measurement differences alter downstream substantive conclusions.

\section{Data and research design}\label{sec:design}

\begin{figure*}[tp]
\centering
\includegraphics[width=\textwidth]{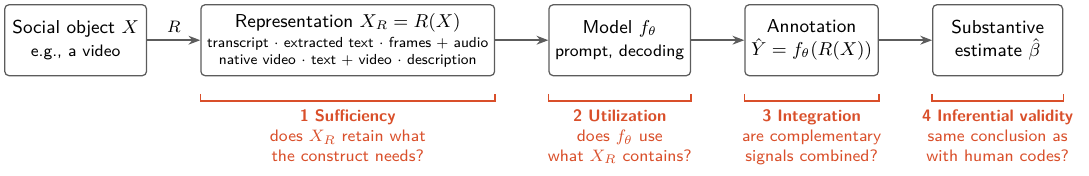}
\caption{Representation as part of measurement. We apply a representation procedure $R$ to the original object $X$, producing the input $X_R$ that the model observes. We then use the model's codes $\hat{Y}$ in substantive analysis. Our four validation checks address different steps of this process.}
\label{fig:framework}
\end{figure*}

In our experiments, we hold the model, codebook prompt, decoding settings, and evaluation target constant while varying how we represent each item (Figure~\ref{fig:framework}). In addition, we provide the same item metadata with every route.

\subsection{Validation datasets}

We use three datasets with human reference labels (Table~\ref{tab:datasets}). They differ in language, domain, and the channels relevant to their constructs.

We use \emph{News on TikTok} as our primary dataset. It contains videos that German-language news outlets in Germany, Austria, and Switzerland posted on TikTok in 2023. Trained coders annotated the videos using a published codebook \citep{Mayeretal2025newstiktok}. We draw a class-quota sample of 603 test videos from the 7{,}978 videos that the published release classifies as news. We sample until each class of four verbally defined constructs (geographic scope, negative news, conflict, and scope) contains about 200 videos. Within each class, we weight candidates by their country's share of the corpus and count each selected video toward every class it matches. We exclude 132 development videos that we drew earlier using the same procedure and used only for format checks. The test draw yields 612 videos, but we could not download nine of them and did not replace them (Appendix~\ref{app:sample}).

This procedure over-represents minority classes of the four quota constructs. For example, the human coders identify conflict in 33\% of test videos, compared with 14\% of the corpus. The other constructs remain close to their corpus shares. We use the sample to evaluate classification and to compare automated and human-coded sample prevalences; we do not estimate prevalence in the corpus or on the platform. Professional news videos often repeat information across voice-over, on-screen headlines, and footage, making them a useful setting for testing how much video adds beyond extracted text.

We use 400 videos from \emph{HateMM}, a dataset of BitChute videos with human hate-speech annotations \citep{Das2023hatemm}: 200 hateful and 200 non-hateful videos. We include hateful videos only if the annotated hateful segment starts within the first 120 seconds, so that every route includes material from the relevant segment. These user-generated videos can convey hate through speech, symbols, imagery, or a combination of channels. The original annotators also recorded which community each video targets or concerns, for hateful and non-hateful videos alike. We analyze two of these target-group labels as separate constructs: whether a video targets Black people (207 of 400 videos) and whether it targets Jewish people (47 of 400). Comparing three constructs on the same videos lets us test whether representation effects differ by construct while holding the material constant.

We use 650 clips from \emph{MUStARD}, a dataset of English-language sitcom utterances with audio and human sarcasm labels \citep{Castro2019mustard}. We reserve 40 of the original 690 clips for development (20 per class), leaving 650 test clips, of which 325 are sarcastic. We include the preceding dialogue that the dataset provides as context. Sarcasm may depend on delivery as well as on words and context. We also use MUStARD's human transcripts to distinguish automatic speech-recognition errors from differences between representations.

\begin{table*}[tp]
\centering\small
\setlength{\tabcolsep}{4pt}
\renewcommand{\arraystretch}{1.15}
\caption{Validation datasets.}
\label{tab:datasets}
\begin{tabularx}{\textwidth}{@{}L{0.16\textwidth}YYY@{}}
\toprule
 & \textbf{News on TikTok}\newline\citet{Mayeretal2025newstiktok} & \textbf{HateMM}\newline\citet{Das2023hatemm} & \textbf{MUStARD}\newline\citet{Castro2019mustard}\\
\midrule
Language and domain & German; professional news from 18 outlets in Germany, Austria, and Switzerland & English; user-generated videos on BitChute & English; TV sitcoms\\
\addlinespace
Test sample & 603 videos & 400 videos & 650 clips\\
\addlinespace
Reference labels & Trained coders; published codebook. Final two-coder pretest: $N=51$, $\alpha=.82$--$1.00$ across the constructs used here & Two annotators plus expert adjudication; $\kappa=.63$ & Human annotations; $\kappa=.23$ and $.59$ in two annotation phases. Human transcripts available\\
\addlinespace
Constructs & Seven verbally defined constructs (news values and platform features); two audiovisual anchors & Hate; targets Black people; targets Jewish people & Sarcasm in the target utterance\\
\addlinespace
Relevant channels & Channels largely repeat the same information; the anchors require picture or sound & Speech, text, symbols, and imagery can convey hate and identify its target & Delivery and context can convey sarcasm\\
\bottomrule
\end{tabularx}
\end{table*}

\subsection{Constructs and their information requirements}

We use the coding definitions to classify channels as required, potentially informative, or unnecessary for each construct (Table~\ref{tab:requirements}). This table is a descriptive classification assembled during interpretation of the results; it is not an independently validated measure of channel requirements. We then check whether representation effects match those information requirements.

The seven news constructs that we call \emph{verbally defined} are those whose definitions refer to what a story is about or to prompts that can be spoken or written. We include geographic scope, negative news, conflict, scope, two kinds of calls to interact, and other people's voice. We select these constructs because they occur often enough in the sample and their definitions refer mainly to verbal content. We examine purely visual and auditory constructs separately. Two of the seven verbally defined constructs also involve nonverbal information. A call to interact on TikTok can be ``purely visual or via an animation,'' and whether ``the voice of people other than the journalist'' is heard depends on who speaks, which a transcript does not record. The two \emph{audiovisual anchors}, visual set-up and music, require visual or auditory evidence and serve as manipulation checks: providing the relevant channel should improve performance.

\begin{table*}[tp]
\centering\footnotesize\setlength{\tabcolsep}{3pt}
\caption{Information requirements by construct, classified from the coding definitions. \req{} required; \pinf{} potentially informative; \nr{} not required. ``Audio'' refers to non-verbal sound and to properties of the voice (who speaks, tone), as distinct from the words spoken.}
\label{tab:requirements}
\begin{tabularx}{\textwidth}{@{}l>{\centering\arraybackslash}p{0.94cm}>{\centering\arraybackslash}p{1.25cm}*{2}{>{\centering\arraybackslash}p{0.94cm}}Y@{}}
\toprule
Construct & Speech & \makecell{On-screen\\text} & Visual & Audio & Basis in the definition\\
\midrule
\multicolumn{6}{@{}l}{\emph{News on TikTok, verbally defined}}\\
Geographic scope (V12) & \pinf & \pinf & \pinf & \nr & location or topic of the event compared with the outlet's nationality (in metadata); explicit or implicit cues\\
Negative news (V15) & \pinf & \pinf & \pinf & \nr & topic of the story; ``code the topic, not the tonality''\\
Conflict (V16) & \pinf & \pinf & \pinf & \nr & focus on controversy, strikes, fights, warfare\\
Scope (V17) & \pinf & \pinf & \nr & \nr & story ``describes'' how many people are involved or affected\\
Call to interact, TikTok (V9b) & \pinf & \pinf & \pinf & \nr & ``visual- or audio-based prompts''; can be ``purely visual or via an animation''\\
Call to interact, off-platform (V9c) & \pinf & \pinf & \pinf & \nr & ``visual- or audio-based prompts'' to leave TikTok\\
Other people's voice (V8b) & \pinf & \pinf & \pinf & \req & whether voices other than the journalist's are used\\
\multicolumn{6}{@{}l}{\emph{News on TikTok, audiovisual anchors}}\\
Visual set-up (V6) & \nr & \nr & \req & \nr & whether a journalist is visible; type of footage\\
Music (V8d) & \nr & \nr & \nr & \req & whether music is used\\
\multicolumn{6}{@{}l}{\emph{Other datasets}}\\
Hate & \pinf & \pinf & \pinf & \pinf & content that ``promotes discrimination or disparages''; channel not specified\\
Targets Black people & \pinf & \pinf & \pinf & \nr & does the video ``target or concern'' the group ``as a community (for example by talking about, depicting, or attacking them)''\\
Targets Jewish people & \pinf & \pinf & \pinf & \nr & as above\\
Sarcasm & \req & \nr & \pinf & \pinf & meaning of the target utterance; signalled by words, context, or ``delivery (tone of voice \ldots{} facial expression)''\\
\bottomrule
\end{tabularx}
\end{table*}

\subsection{Models}

We use six open-weight LLMs that accept both images and audio and return text. They are Gemma~4 12B and Gemma~4 E4B \citep{Google2026gemma4}, Qwen3-Omni 30B-A3B \citep{Xuetal2025qwen3omni}, Qwen2.5-Omni 7B \citep{Xuetal2025qwen25omni}, Nemotron-3 Nano Omni 30B-A3B \citep{NVIDIA2026nemotronomni}, and MiniCPM-o~4.5 \citep{Yaoetal2024minicpmv}. The set includes dense and mixture-of-experts architectures with different approaches to audio processing. We require audio input so that every model can receive speech audio in the frames + audio route. We exclude models without audio input (for example Gemma~4 26B and 31B) because they cannot run that route as defined. We list exact checkpoints and revisions in Appendix~\ref{app:models}.

We use the six models to assess whether representation effects recur across architectures, sizes, and developers. Our focus is on these effects and their variation across models, rather than on a model ranking.

We run all models on one NVIDIA A100 (80\,GB) with vLLM~0.30 \citep{Kwonetal2023vllm}, using bfloat16 precision and greedy decoding. We disable reasoning (``thinking'') modes and use a JSON-schema grammar to restrict outputs to each construct's legal codes. Completed coding calls return parseable, in-scale answers; this does not prevent input-processing failures. We code the nine news constructs together and the two HateMM target variables together. Hate and sarcasm each use a separate single-construct call. We disable prefix caching and the multimodal preprocessing cache, except that the capped full-video runs use the preprocessing cache (Appendix~\ref{app:models}).

\subsection{Representation routes}

We compare the nine routes in Table~\ref{tab:routes}, using each model to code every item through every route that its dataset supports. We provide at most the first 120 seconds of each video, except in the full-video route, where we supply the original file. We also provide the same metadata across routes: outlet and nationality, caption, and hashtags for news; show, speaker, and preceding dialogue as a human transcript for sarcasm. HateMM provides no metadata. We use the same prompt across routes within each corpus, changing only one sentence to describe the input material (Appendix~\ref{app:prompts}). The news and hate prompts use definitions from the source coding guidelines; for sarcasm and the HateMM target variables, we supply operational definitions consistent with the source annotations. We fix each prompt before running its corresponding comparison and retain it across models and routes. The original news, hate, and sarcasm prompts were frozen before the first test-set calls; the target-group prompts and additional video routes were introduced later. We use the development sets for format and input-processing checks, not accuracy tuning.

\begin{table*}[tp]
\centering\small
\caption{Representation routes. ASR: automatic speech recognition (Whisper large-v3-turbo). OCR: optical character recognition at one frame per second, near-duplicate lines removed. The three video routes differ in who decides which pictures the model sees and in how pictures and sound are packaged.}
\label{tab:routes}
\begin{tabularx}{\textwidth}{@{}L{2.2cm}YL{5.4cm}@{}}
\toprule
Route & What the model receives & Notes\\
\midrule
\multicolumn{3}{@{}l}{\emph{Text routes: we convert the video into text; the model reads only text}}\\
Metadata only & caption, hashtags, outlet (news); show, speaker, preceding dialogue (sarcasm) & floor; the metadata are also part of every other route\\
Transcript & + ASR text & extraction run once, reused by all models\\
Extracted text & + ASR text + OCR text & sarcasm: identical to transcript (no on-screen text)\\
Human transcript & + human transcript & sarcasm only; isolates ASR error\\
Describe first & + a description of the video written by the same model & first call: the model watches frames + audio and describes speech, on-screen text, visuals, and sound; second call: coding from that text\\
\addlinespace
\multicolumn{3}{@{}l}{\emph{Video routes: the model sees pictures and hears sound itself}}\\
Frames + audio & 16 still images that we extract (448\,px, evenly spaced over the first 120\,s) and the audio as separate 30-s clips & each image is encoded on its own, like a photo; images and sound are not linked in time; identical pixels and audio for every model\\
Native video & one video file that we prepare (first 120\,s, 448\,px); the model's own video pathway samples 16 frames from it & frames encoded as a video sequence by the model's video encoder; audio aligned with the frames in time in Qwen2.5-Omni and Nemotron, separate clips otherwise$^a$\\
Full video & the original file, untouched: whole length, original resolution & the model's video pathway decides how many frames to take and at what resolution; capped fallback tested where the original does not fit$^b$\\
Text + video & frames + audio and extracted text in one call & both inputs together\\
\bottomrule
\multicolumn{3}{@{}p{\textwidth}@{}}{\footnotesize $^a$For Qwen3-Omni, vLLM~0.30 failed on time-aligned audio in the video pathway; its native-video route therefore received the audio as separate clips (Appendix~\ref{app:models}). $^b$The whole file sampled to 32 frames at the model's default resolution, with the audio as 30-s clips of the first 120 seconds; tested as a fallback when the original fails for more than 5\% of items. No capped fallback met the 95\% completion threshold where the original failed it (Appendix~\ref{app:videoinput}).}
\end{tabularx}
\end{table*}

The video routes differ in sampling and encoding. For \emph{frames + audio}, we supply sixteen still images and separate audio clips. The model encodes each image independently, without explicit alignment to the sound. This input is identical across models and also supplies the video component of text + video.

For \emph{native video}, we supply a file containing the first 120 seconds at 448 pixels. Each model's video pathway samples sixteen frames and encodes them as a sequence. Some models encode these frames more compactly than independent images; two align the audio with the frames. This route keeps the visual budget approximately constant while allowing architecture-specific processing.

For \emph{full video}, we supply the original file and use the model's default frame sampling and resolution. Long or high-resolution files may exceed context or memory limits. We also test a capped version with 32 frames from the full file and audio clips from the first 120 seconds. We retain a full-video cell only if at least 95\% of items are processed. All retained cells use the untouched original; no capped fallback rescued an otherwise excluded cell. Appendix~\ref{app:videoinput} reports completion counts, including the 20-video capped-news pilots for the two Qwen models.

\subsection{Evaluation}\label{sec:eval}

We evaluate each construct against the human reference labels using macro-F1, the unweighted mean of F1 across the codebook's categories. We exclude missing or unclear ($-99$) human labels separately for each construct. Among the remaining items, we count missing or invalid model outputs as incorrect, including failures caused by inputs that exceed a model's limits. To summarize a group of constructs, we average their macro-F1 scores rather than pooling their items. We also report nominal Krippendorff's $\alpha$ against the human labels \citep{HayesKrippendorff2007answering}.

We report macro-F1 for every route, construct, and model, with 95\% confidence intervals from an item bootstrap (1{,}000 resamples). We compare routes within each model and construct, using the same items. For each model and construct we identify the observed best route and compare every other route with it through the paired bootstrap distribution of the difference in macro-F1. Using a margin of $\pm.05$ macro-F1, we call a route \emph{practically equivalent} to the best if the 90\% interval of the difference lies within $\pm.05$, \emph{worse} if the interval lies entirely below $-.05$, and \emph{inconclusive} otherwise. Because we select the comparator for having the highest observed score, the comparison favors it. We therefore treat this as a descriptive diagnostic rather than a confirmatory equivalence test. We also report margins of .03 and .10 as sensitivity checks.

We summarize the loss from choosing a route as its \emph{shortfall}, the distance in macro-F1 between that route and the best route of the same row. We decompose the variation of construct-level macro-F1 across models, routes, and constructs into main effects and interactions; with one value per cell, this is an exact descriptive decomposition, not a test. Where we compare two specific routes, we report the paired bootstrap difference (2{,}000 resamples, same items for both routes) with its 95\% interval. We do not test differences across the six models, which are not independent observations.

We test audio use by coding the same visual input with and without its audio track. We run this \emph{audio ablation} on a random subsample of 150 news videos, where we also vary the number of frames, on all 400 HateMM videos, and on all 650 sarcasm clips. We run additional diagnostic tests for the two Gemma models because their initial news results show little response to audio. We test an audio-only route, the media placement recommended in the Gemma~4 documentation (images before the instruction, audio after it), a direct question about music, and transcription with the developer's recommended prompt. We also repeat the tests with a second inference backend, Hugging Face transformers.

To assess \emph{integration}, we classify each model's item $\times$ construct decisions into four cases: both routes are correct, only extracted text is correct, only frames + audio is correct, or both are wrong. We then calculate an oracle score by selecting the correct label whenever either route provides it and the extracted-text label when both routes are wrong. This diagnostic uses the human reference labels to show the benefit of resolving disagreements correctly; we cannot use it to code unlabeled data.

To assess \emph{inferential validity}, we re-estimate an engagement regression on the news data, replacing the human codes with each route's codes (Section~\ref{sec:infer}). For each coefficient, we examine whether pipelines with higher macro-F1 produce estimates closer to the human-code estimate. We also check whether selecting the highest-F1 route selects the route with the smallest coefficient error. For every binary construct across the three datasets, we compare automated and human-coded sample prevalences, both across all routes and after selecting each model's highest-F1 route.

We assess reproducibility by re-running 50 news videos on six routes with identical settings.

An initial analysis plan, the original codebook prompts, and the inference code were frozen on 10 September 2026, with their SHA-256 hashes recorded locally; the plan was not deposited in a public registry. After interim results, we moved from laptop inference to the A100 and reran the study, expanded the model set, and added the native- and full-video routes and HateMM target-group comparisons. The broader grid, integration, prevalence, and downstream route-selection analyses developed during this extension. We therefore report these comparisons as descriptive and exploratory, while retaining the initial plan's $\pm.05$ equivalence margin. Appendix~\ref{app:models} gives further details. We used generative AI in preparing replication material and supplementary information, as specified in the declarations.

\section{Results}\label{sec:results}

\begin{figure*}[p]
\centering
\includegraphics[width=\textwidth,height=.79\textheight,keepaspectratio]{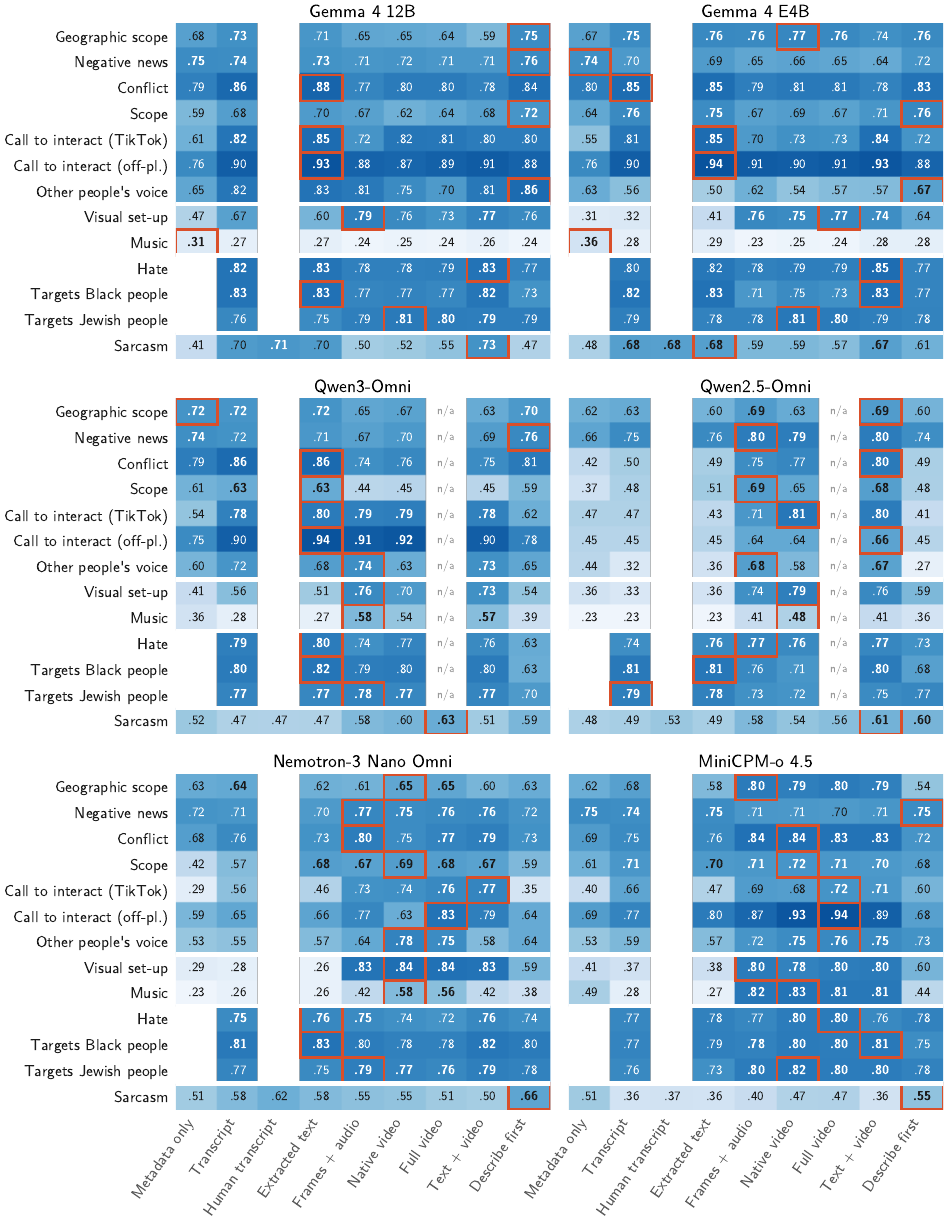}
\caption{Macro-F1 for every route, construct, and model. Rows: the nine news constructs (verbal constructs above the first white line, the two audiovisual anchors below), hate, the two HateMM target groups, and sarcasm. Orange frame: observed best route in the row. Bold: best or practically equivalent to it (90\% interval of the paired difference within $\pm.05$). n/a: the original-file route failed the 95\% completion threshold and no tested capped fallback met it. Empty: route not defined for the dataset (no metadata for HateMM; human transcripts only for sarcasm). For sarcasm, extracted text equals the transcript.}
\label{fig:grid}
\end{figure*}

Completed calls on extracted text, frames + audio, and native video produced no unparseable or out-of-codebook answers. Native-video processing nevertheless failed on 12 of 400 HateMM videos for Qwen2.5-Omni and on 3 for Nemotron, in both the hate and target-group calls; these failures count as incorrect. Extracted text and frames + audio completed on all test items. Figure~\ref{fig:grid} shows macro-F1 for every route, construct, and model on all three datasets. We summarize scores by construct group in Appendix Table~\ref{tab:fullgrid} and provide all cell values and intervals in the replication package.

\subsection{Representation sensitivity}\label{sec:rq1}

For news, we compare representations on the same videos and constructs with constant prompts and models. On the seven verbally defined constructs, native-video macro-F1 falls in a narrow range from .70 to .77 across all six models. Extracted-text performance spans .52 to .80. The direction of the difference therefore reverses across models. Gemma~4 12B, Gemma~4 E4B, and Qwen3-Omni perform better from extracted text, whereas Qwen2.5-Omni, Nemotron, and MiniCPM-o perform better from native video. The paired native-video minus extracted-text differences range from $-.06$ to $+.18$, with all intervals excluding zero (Appendix~\ref{app:contrasts}).

\begin{figure*}[tp]
\centering
\includegraphics[width=\textwidth]{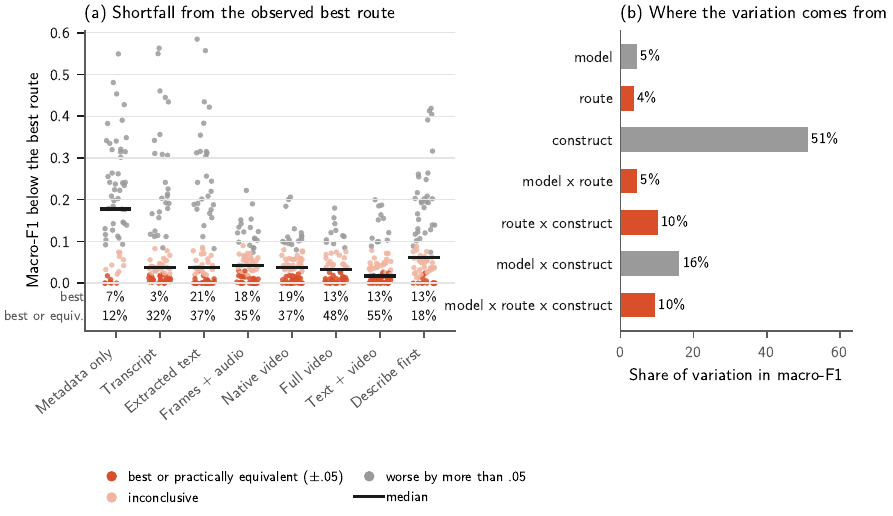}
\caption{(a) Shortfall of each route from the observed best route of the same model and construct (78 combinations; full video 54, metadata only 60). Orange: best or practically equivalent (90\% interval of the paired difference within $\pm.05$); light orange: inconclusive; grey: worse by more than .05. Numbers below: share of combinations in which the route is best, and in which it is best or practically equivalent. (b) Decomposition of the variation in construct-level macro-F1 across six models, six routes available for all constructs (transcript, extracted text, frames + audio, native video, text + video, describe first), and 13 constructs.}
\label{fig:routechoice}
\end{figure*}

This variation also appears across the full grid. Across the 78 model--construct combinations, no route dominates (Figure~\ref{fig:grid}). Extracted text is the observed best route in 16 combinations, native video in 15, frames + audio in 14, text + video and describe first in 10 each, full video in 7, metadata only in 4, and the transcript in 2. Relative to the observed best route, 24\% of the 510 pairwise comparisons are practically equivalent within the pre-specified $\pm.05$ margin, 37\% are worse by more than .05, and 38\% remain inconclusive (Figure~\ref{fig:routechoice}a; Appendix Table~\ref{tab:equivalence}). The typical shortfall is small for several routes. The median is .02 for text + video and .03--.04 for the main text and video routes. Every route nevertheless has combinations on which its shortfall is substantial. The size of the best-or-equivalent set changes with the margin; many comparisons remain inconclusive.

The variance decomposition shows that representation effects depend on the model and construct (Figure~\ref{fig:routechoice}b). Differences between constructs account for 51\% of the observed variation in macro-F1. The route main effect accounts for 4\% and the model main effect for 5\%, while route $\times$ construct accounts for 10\%, model $\times$ route for 5\%, model $\times$ construct for 16\%, and the three-way interaction for another 10\%. Most variation associated with representation occurs through its interactions with the construct and model.

The ordering of routes also varies in HateMM and MUStARD. For hate, extracted text (.76 to .83) or text + video (.76 to .85) is the observed best route for four of six models, and no video route improves on extracted text by more than .02. For sarcasm, the preferred route changes sharply across models. The Gemma models do best from the transcript or text + video, Qwen3-Omni from the full clip, Qwen2.5-Omni from text + video, and Nemotron and MiniCPM-o from a model-written description. When models can process the untouched file, its pooled performance stays within .04 of native video. Providing the original file offers no consistent advantage over the reduced representations.

\subsection{Construct dependence}\label{sec:rq2}

We compare constructs on the same items to hold language, genre, sample, and annotation source constant. On the 603 news videos, frames + audio improves on extracted text by .36 macro-F1 on average for visual setup, with the interval above zero in all six models. For other people's voice, the mean gain is .12 and five of six intervals lie above zero. By contrast, the six remaining verbal constructs range from a mean difference of $-.02$ to $+.08$, with models disagreeing in sign (Appendix~\ref{app:contrasts}). For music, the audiovisual advantage varies with the model's use of audio. Relative to the verbal-construct mean, the additional gain for music ranges from $-.02$ to $+.45$ with frames + audio and is positive with a confidence interval above zero for five of six models under native video.

We make a second same-item comparison using HateMM. On the same 400 videos, all six models code the Black-target variable best from extracted text or text + video, while five of six code the Jewish-target variable best from a video route (Figure~\ref{fig:grid}). Frames + audio gains more over extracted text for the Jewish-target variable than for the Black-target variable in every model; the difference is .04 to .12 in five models, with the paired interval above zero in four. Thus, the preferred representation can change with the target construct even when we hold the videos and models constant.

The wider comparison follows a similar pattern. Video performs best for visual setup in every model and for music in the four models that use its audio signal. Results for the verbal news constructs vary by model. We interpret cross-dataset contrasts cautiously because language, genre, construct, and annotation source vary together. The same-item comparisons provide clearer evidence that representation effects depend on both the information a construct requires and the model's ability to use it.

\subsection{Modality utilization}\label{sec:rq3}

\begin{figure*}[tp]
\centering
\includegraphics[width=\textwidth]{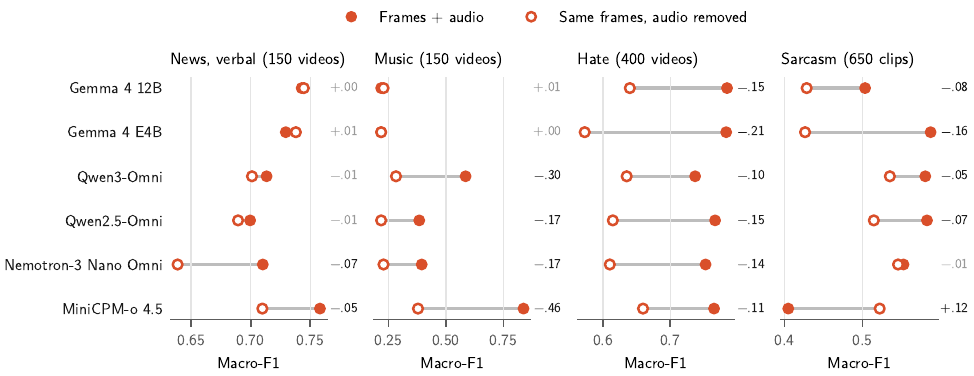}
\caption{Audio ablation: macro-F1 of the frames + audio input with and without the audio track. News: random subsample of 150 videos (seven verbal constructs pooled; music). Hate: 400 videos (not run for the two target groups). Sarcasm: 650 clips. Numbers give the change when the audio is removed.}
\label{fig:ablation}
\end{figure*}

We test models' use of audio by removing it from the frames + audio route while keeping the visual input constant (Figure~\ref{fig:ablation}).

The effect of removing audio varies by task. For hate, removing the soundtrack lowers macro-F1 in every model by .10 to .21. For sarcasm, it costs Gemma~4 12B .08, Gemma~4 E4B .16, and Qwen2.5-Omni .07. Music detection drops by .17 to .46 in the four non-Gemma models. Yet on the seven verbal news constructs, removing audio changes the two Gemma models by .01 or less, while it costs Nemotron .07 and MiniCPM-o .05. These differences show why researchers need to assess audio use for each task.

Further tests distinguish speech transcription from music detection in the Gemma models (Table~\ref{tab:gemmadiag}). Both Gemma models transcribe English speech well when we use the developer-recommended transcription instruction. Their median word error rate is .11 on the sitcom clips. An earlier generic prompt had made Gemma~4 12B appear unable to transcribe many clips, so that result was a prompting artifact. Media placement affects the audio-only news task. For Gemma~4 12B, coding the news constructs from audio alone rises from .69 when audio precedes the instruction to .76 when it follows it, as the documentation recommends; once frames are present, the change is only .01. Neither Gemma model reliably detects music. When we ask directly whether an audio clip contains music, the Gemma models answer yes for only 0--2\% of the first 30-second audio clips. The human labels identify music in 72\% of the corresponding whole videos, but we do not have separate reference labels for these short clips. This difference in observation windows limits the direct prevalence comparison. We find the same pattern when we repeat the probe with Hugging Face transformers, making a vLLM-specific explanation unlikely. The models can process speech, but this ability does not extend reliably to music detection in these tests.

\begin{table*}[tp]
\centering\footnotesize
\caption{Gemma audio diagnosis on 150 news videos, with Nemotron as control. Placement ``after the instruction'' follows the Gemma~4 documentation (images before the text, audio after it); our main runs placed all media before the text. The direct music probe uses the first 30 seconds; 72\% of the corresponding whole videos, not necessarily these clips, have a positive human music label.}
\label{tab:gemmadiag}
\fitwidth{
\begin{tabular}{@{}lccc|ccc@{}}
\toprule
 & \multicolumn{3}{c|}{Seven verbal constructs} & \multicolumn{3}{c}{Music: share coded as present}\\
Input (150 news videos) & Gemma 12B & Gemma E4B & Nemotron & Gemma 12B & Gemma E4B & Nemotron\\
\midrule
Metadata only & .67 & .66 & .52 & 8\% & 20\% & 0\%\\
Extracted text & .80 & .75 & .62 & 2\% & 3\% & 1\%\\
Frames only (audio removed) & .74 & .74 & .64 & 1\% & 0\% & 1\%\\
Frames + audio, audio before the instruction & .74 & .73 & .71 & 0\% & 0\% & 14\%\\
Frames + audio, audio after the instruction & .75 & .75 & .71 & 7\% & 3\% & 14\%\\
Audio only, audio before the instruction & .69 & .69 & .71 & 10\% & 9\% & 17\%\\
Audio only, audio after the instruction & .76 & .72 & .70 & 15\% & 9\% & 19\%\\
\midrule
\multicolumn{7}{@{}l}{\emph{Direct question on audio only: ``Is there music?'' (first 30 seconds); share answering yes}}\\
vLLM, audio after the question & 1\% & 2\% & 27\% & \multicolumn{3}{l}{transformers: 1\% / 2\%}\\
vLLM, audio before the question & 0\% & 2\% & 27\% & \multicolumn{3}{l}{transformers: 0\% / 3\%}\\
\midrule
\multicolumn{7}{@{}l}{\emph{Transcription with the developer-recommended prompt: median word error rate}}\\
English (100 sitcom clips, human reference) & .11 & .11 & .05 & & & \\
German (150 news clips, Whisper reference) & .42 & .45 & .77 & & & \\
\bottomrule
\end{tabular}
}
\end{table*}

All six models transcribe the English sitcom clips well with their recommended prompts (median word error rates .05 to .11). Against the Whisper reference on the German news clips they are substantially weaker (.37 to .77), which helps explain why a dedicated speech-recognition pipeline can be competitive for the German verbal constructs. Frame count matters much less in this application. Relative to 16 frames, 32 changes pooled news macro-F1 by at most .01, while reducing the input to four frames costs up to .04 (Appendix Table~\ref{tab:ablation}).

\subsection{Multimodal integration}\label{sec:integration}

\begin{figure*}[tp]
\centering
\includegraphics[width=\textwidth]{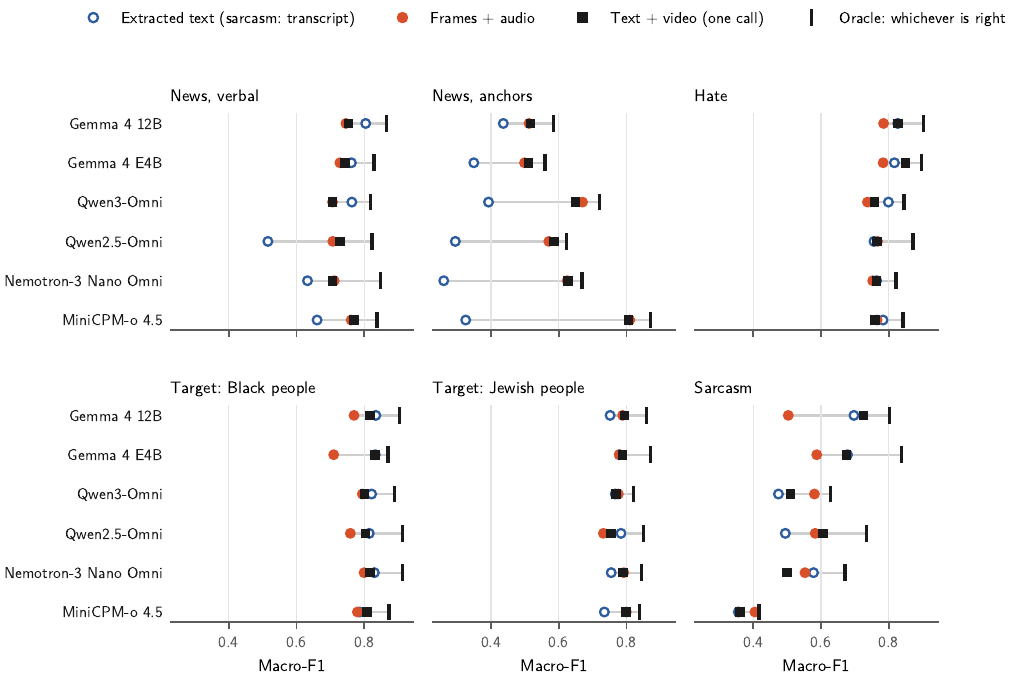}
\caption{Integration across datasets: macro-F1 of extracted text (sarcasm: transcript), frames + audio, both in one call (text + video), and an oracle that selects the correct single-route prediction whenever one is available. The oracle uses human reference labels and is a diagnostic benchmark.}
\label{fig:integration}
\end{figure*}

Extracted text and audiovisual input make complementary errors. Across every model and construct group, some decisions are correct only from extracted text and others only from frames + audio (Appendix Table~\ref{tab:integrationall}). Depending on the model and dataset, 1--24\% of decisions are correct only from text and 3--47\% only from video. A diagnostic oracle that takes the correct answer whenever either route has it would improve on the better single route by .01 to .16 macro-F1, with a median gain of .07 (Figure~\ref{fig:integration}). The two sets of predictions thus contain correct labels that neither route consistently recovers alone.

Providing both representations in one call rarely improves on the better individual route. Across the 36 model--group combinations, text + video ranges from .08 below to .03 above the better single route, with a median difference of $-.003$; it exceeds the better route by .02 or more in only 14\% of cells. Its performance often closely follows one of the component routes. On the news constructs it tracks the video route even for the three models whose extracted-text route is better. Gemma~4 12B, for example, scores .80 from extracted text, .75 from frames + audio, and .75 from both. For the Black-target variable in HateMM, the combined route tracks the text route in five of six models. The combined-input results fall short of the gains suggested by the complementary predictions.

\subsection{Inferential consequences}\label{sec:infer}

We examine whether differences in the automated labels change two quantities: engagement-regression coefficients in the news data and the prevalence of each binary construct in the validation samples.

We re-estimate an OLS model of log(1 + views) on the six binary focal constructs and geographic scope, with outlet fixed effects and HC1 standard errors, using the 599 news videos with complete human codes and view counts. With human codes, negative news ($\hat\beta=.63$, SE .29) and conflict ($\hat\beta=.79$, SE .30) are positively associated with views, while foreign news is negatively associated ($\hat\beta=-.71$, SE .36). We then replace the human codes with each model's codes from extracted text, frames + audio, and native video.

\begin{figure*}[tp]
\centering
\includegraphics[width=\textwidth]{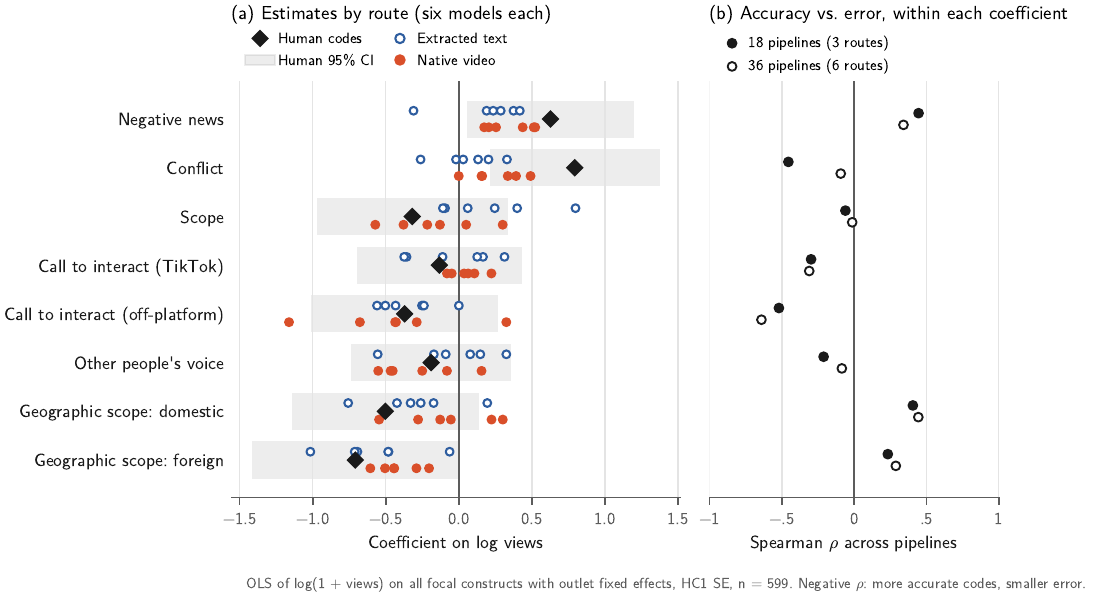}
\caption{Inferential validity in the news data ($n=599$). (a) Coefficients from the engagement regression estimated with human codes (diamond, shaded 95\% interval) and with each model's codes from extracted text and native video. (b) For each coefficient separately, the rank correlation across pipelines between the construct's macro-F1 and the absolute deviation of the coefficient from the human-code estimate. Filled: 18 pipelines (six models $\times$ extracted text, frames + audio, native video); open: 36 pipelines (all six routes used in the regression). Negative values mean that more accurate codes recover the estimate better.}
\label{fig:infer}
\end{figure*}

Using automated codes changes the coefficient estimates (Figure~\ref{fig:infer}a). All 18 automated negative-news coefficients are smaller than the human-code coefficient, with a mean ratio of .49. All 18 conflict coefficients are smaller as well, with a mean ratio of .20. Across all eight coefficients, sign agreement with the human-code estimate is 69\% for extracted text, 73\% for frames + audio, and 77\% for native video (Table~\ref{tab:infer}).

\begin{table*}[tp]
\centering\small
\caption{Recovery of the human-code regression estimates by route (48 estimates per route: 8 coefficients $\times$ 6 models). The human estimates' 95\% intervals are wide (half-widths .55 to .71), so interval overlap is uninformative here.}
\label{tab:infer}
\fitwidth{
\begin{tabular}{@{}lccccc@{}}
\toprule
Route & Estimates & Mean $|\hat\beta_{auto}-\hat\beta_{human}|$ & Same sign & 95\% CI covers $\hat\beta_{human}$ & CI overlap\\
\midrule
Extracted text & 48 & 0.36 & 69\% & 79\% & 100\%\\
Frames + audio & 48 & 0.38 & 73\% & 81\% & 100\%\\
Native video & 48 & 0.32 & 77\% & 90\% & 100\%\\
\bottomrule
\end{tabular}
}
\end{table*}

Selecting a route by macro-F1 does not reliably minimize coefficient error. For each coefficient, we compare the same estimate across 18 pipelines. The Spearman correlation between construct-level macro-F1 and absolute coefficient error ranges from $-.52$ to $+.44$, with a median of $-.14$ (Figure~\ref{fig:infer}b; Appendix Table~\ref{tab:inferestimand}). For three of the eight coefficients, the correlation is positive: pipelines with higher F1 have larger coefficient errors. The model--construct comparisons show the same pattern. Among extracted text, frames + audio, and native video, the route with the highest macro-F1 is also the route with the smallest coefficient error in 17 of 42 comparisons. Uniform random choice among three routes would yield 14 in expectation. Choosing by F1 adds a median error of .08 log points, or .28 human standard errors, relative to the best route; at the 90th percentile it exceeds 1.2 standard errors. These comparisons share items, models, and routes, so we use them descriptively rather than as independent hypothesis tests.

\begin{figure*}[tp]
\centering
\includegraphics[width=\textwidth]{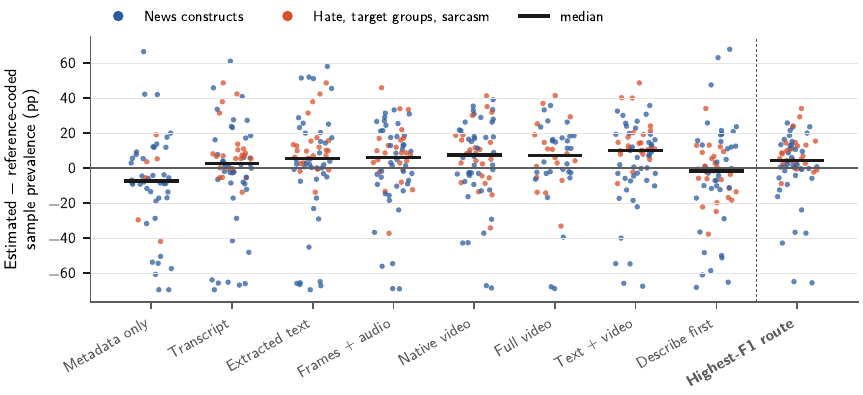}
\caption{Prevalence estimated from automated codes minus the reference-coded prevalence in the sample, for every binary construct, model, and route (496 estimates). Last column: for each model and construct, the deployable route with the highest macro-F1 (66 estimates). Blue: news constructs; orange: hate, target groups, and sarcasm.}
\label{fig:prevalence}
\end{figure*}

We next compare, for every binary construct, the share coded positive by the automated pipeline with the reference-coded share in the same sample (Figure~\ref{fig:prevalence}). Because all three validation samples are balanced or quota samples, these are not estimates of population or platform prevalence. They test whether the automated variable reproduces the prevalence of the human-coded variable on the same items. Across 496 route--model--construct estimates, 59\% differ from the reference by more than 10 percentage points and 32\% by more than 20. Removing metadata-only or restricting the comparison to deployable routes leaves the result essentially unchanged (Appendix Table~\ref{tab:prevrobust}). Even when we select each model's highest-F1 deployable route for each construct, the median absolute error is 9 points; 45\% of estimates remain more than 10 points away and 21\% more than 20 points away. Selecting the most accurate labels therefore improves prevalence recovery but does not ensure it.

\subsection{Robustness and reproducibility}\label{sec:robust}

Four of the six models could process the untouched original files across all three datasets (Nemotron failed on 4.5\% of HateMM videos, mostly because the audio exceeded a size or duration limit). For both Qwen models, most news and HateMM files crashed the vision encoder (Qwen3-Omni) or exceeded the context window (Qwen2.5-Omni, 32k tokens). Capping the visual input at 32 frames did not produce a usable fallback: Qwen2.5-Omni processed 12 of 20 attempted news videos and 147 of 400 HateMM videos; Qwen3-Omni processed none of the 20 news videos and only one HateMM video. The news pilots do not establish completion rates for all 603 videos (Appendix Table~\ref{tab:feasibility}). Uploading the original file is therefore infeasible for some models and datasets. Where full video ran, it performed like native video (within .04 pooled).

The standardized frames + audio input and the model's own video pathway give the same pooled result on the news constructs (differences within $\pm.012$), although individual constructs can move by up to $\pm.15$; frames at 896 instead of 448 pixels change pooled macro-F1 by at most .02 (Appendix~\ref{app:videoinput}).

The agreement results give a less favorable assessment than the macro-F1 values, which reach .94. In Krippendorff's $\alpha$, only 12\% of the 252 results from six models, six routes (metadata only, transcript, extracted text, frames + audio, text + video, describe first), and seven news constructs reach .667, and only for one construct does the best route reach the agreement of the two trained coders in the published pretest. When we treat the six models as six coders, their agreement ranges from $\alpha=.27$ to .70 (Appendix~\ref{app:alpha}).

We re-run 50 news videos on six routes with identical settings on the same GPU. Models reproduce 97.3\% to 99.8\% of individual codes and all nine codes for 81\% to 98\% of item--route pairs (Appendix~\ref{app:rerun}). Batched GPU inference is not fully deterministic. The least stable here were the two mixture-of-experts models (97.3\% and 97.9\%), with one dense model close behind (98.3\%); six models do not support a conclusion about architectures, and no route was consistently less stable than the others.

\subsubsection{Cost}
Coding from extracted text required 0.09 to 0.26 A100 GPU-seconds per news video, frames + audio 0.36 to 1.11 seconds (Appendix~\ref{app:cost}). At the University of Zurich computing rate of CHF 0.081 per A100-hour, coding 1{,}000 news videos therefore costs CHF 0.002 to 0.006 from extracted text and CHF 0.008 to 0.025 from frames + audio. At on-demand prices of commercial cloud providers, US\$3.43 (Amazon Web Services) to US\$3.67 (Microsoft Azure) per A100-hour \citep{AWS2026ec2pricing,Microsoft2026azurepricing}, the same jobs cost US\$0.09 to 0.27 and US\$0.34 to 1.13. For these inference runs, the absolute cost differences between representations are small.

\FloatBarrier
\section{Discussion}\label{sec:discussion}

Our study shows that input representation belongs in the specification of a video annotation instrument. Transcripts, sampled frames with audio, native video, and original files expose different information to the model. Switching between them can change the resulting variables while the model, prompt, decoding settings, and source videos remain fixed.

For the seven verbal news constructs, native-video performance is similar across models (.70--.77 macro-F1), while extracted-text performance ranges from .52 to .80 and changes their ordering. Across the full comparison, the main effect of representation is small relative to its interactions with model and construct. A representation that works well in one model--construct combination may perform poorly in another. Reporting the model name and prompt is therefore insufficient for reproducing a multimodal measure. Researchers should also document speech and OCR extraction, frame counts and resolution, audio alignment, truncation, and media placement within the prompt. Each affects the evidence the model receives.

Our findings do not establish a general preference for text or video. Extracted text can outperform audiovisual input on verbal news constructs, whereas visual setup requires pictures. On the same HateMM videos, audiovisual input has different advantages for the Black-target and Jewish-target variables. The codebook helps identify relevant channels, but performance also depends on whether the model uses them. Full video adds a feasibility constraint. Where models can process the original file, performance is close to that of the controlled video routes. Two models, however, fail to process most original news and HateMM files, and the capped trials do not establish a reliable fallback. Researchers need to test whether their model can process a candidate representation reliably before scaling up annotation.

An audio-capable model may use sound for one task and largely ignore it for another. The ablations show large effects for hate, smaller effects for some sarcasm models, and little effect for several models on verbal news content. Both Gemma models transcribe English speech with the recommended prompt but rarely detect music. Changing media placement mainly affects the audio-only task, and the music failure recurs with a second inference backend. These findings support testing modality use for the specific construct and input configuration. Integration also requires a separate check. The oracle gains a median .07 macro-F1 by selecting correct predictions from the two routes using the reference labels. The combined-input route exceeds the better component by at least .02 in only 14\% of model--group combinations. Receiving several modalities therefore does not establish that a model combines their useful signals.

The downstream analyses show why annotation validation should include the intended estimate. Automated codes substantially attenuate the negative-news and conflict coefficients, and higher macro-F1 does not consistently correspond to smaller coefficient error. Among the three main routes, selecting the highest-F1 route also selects the closest coefficient estimate in 17 of 42 model--construct comparisons. The prevalence analysis yields a similar result. Even after selecting the highest-F1 deployable route for each model and construct, 45\% of estimates differ from the human-coded sample prevalence by more than ten percentage points. This is consistent with prior work on misclassification bias \citep{TeBlunthuisetal2024misclassification}: the effect of coding errors depends on their distribution across the data. F1 and agreement statistics remain useful assessments of a classifier, but neither alone establishes whether its labels support the intended estimate. A human-coded subset can support both forms of validation. Researchers can use it to compare representations before scaling up and to assess or correct error in downstream estimates \citep{Egamietal2023surrogates}. Similar checks may be useful beyond video wherever preprocessing changes the evidence available to a computational measure.

Based on our findings, we recommend five checks for multimodal LLM annotation.
\begin{enumerate}
\item \emph{Validate more than one plausible representation.} Use a human-coded subset to compare model--representation combinations that researchers could reasonably deploy, and use a substantively defined equivalence margin rather than treating non-significance as interchangeability.
\item \emph{Match the candidate representations to the construct, then test utilization.} The codebook indicates which channels could matter; a simple modality ablation shows whether the model actually uses them on the task.
\item \emph{Do not assume that additional modalities are integrated.} Compare a combined route with its components. Check whether the combined input improves on the better component when the individual routes make different errors.
\item \emph{Validate the downstream quantity when automated labels feed substantive inference.} Classification accuracy and estimand recovery need not select the same pipeline. Where possible, use the human-coded subset to quantify or correct the resulting measurement error.
\item \emph{Archive the representation as part of the instrument.} Record extraction tools, frame and audio handling, prompt placement, model checkpoints, outputs, and a re-run check. The representation and the realized model outputs are part of what must be reproducible.
\end{enumerate}
These checks should include model-specific input conventions and processing limits. For the inference runs measured here, GPU cost differences are small enough to give priority to validation among feasible routes.

The evidence comes from three datasets and six open-weight models that fit on one A100 GPU. The primary application is German news, while HateMM and MUStARD are English; larger proprietary systems or future checkpoints may use modalities differently. The regression test is available only for the news data, although the prevalence analysis spans all three datasets. ``Native video'' is also not one perfectly standardized representation because each architecture implements its video pathway differently, and Qwen3-Omni required audio to be supplied separately. The main frame-based routes see at most the first 120 seconds and sixteen frames; more frames add little here, but faster-cut or much longer material may behave differently.

The validation samples also set boundaries on the claims. The news sample deliberately over-represents minority classes of four quota constructs, while HateMM and MUStARD are balanced samples. The prevalence analysis therefore concerns recovery of the human-coded prevalence in those samples, not population prevalence. Human labels are references rather than error-free ground truth. The news release is single-coded outside its published reliability pretest, and the hate and sarcasm datasets report imperfect inter-annotator agreement. Finally, comparisons across datasets confound the construct with language, genre, and annotation; we therefore treat the same-item contrasts as the strongest evidence for construct dependence.

Overall, across six open-weight models and three datasets, the choice of input representation changes video annotation quality, with effects that depend on the model and construct. Audio availability does not ensure that a model uses sound for the task, and combined text and video inputs rarely improve on the better individual route. Higher annotation accuracy does not reliably yield regression coefficients or sample prevalences closer to the human-coded estimates. Researchers using multimodal LLMs should validate the representation alongside the model and prompt, and assess the estimates they intend to derive from the labels.

\backmatter

\section*{Statements and Declarations}

\bmhead{Funding}
We received no funding for this study.

\bmhead{Competing interests}
We have no competing interests relevant to this study to declare.

\bmhead{Ethics approval and consent to participate}
Not applicable.

\bmhead{Consent for publication}
Not applicable.

\bmhead{Data availability}
We will deposit the prompts, codebooks, and all model outputs on the Open Science Framework (OSF) upon publication. The three datasets are available from their original publishers: News on TikTok \citep{Wedeletal2025dataset,Mayeretal2025newstiktok} from the GESIS Data Archive, \url{https://doi.org/10.7802/2863}; HateMM \citep{Das2023hatemm} from Zenodo, \url{https://doi.org/10.5281/zenodo.7799469}; and MUStARD \citep{Castro2019mustard} from GitHub, \url{https://github.com/soujanyaporia/MUStARD}.

\bmhead{Code availability}
We will deposit all code needed to reproduce the analyses, from data preparation and inference to every table and figure, on the Open Science Framework (OSF) upon publication.

\bmhead{Use of generative AI}
We used Claude Sonnet 5 (Anthropic) for language editing during manuscript preparation and Claude Code Opus 5.5 (Anthropic) to prepare the replication material and organize the supplementary information. We designed the study, analyzed the data, and take full responsibility for the content.

\bmhead{Author contributions}
C.P. conceived and designed the study, collected and prepared the data, implemented the annotation pipelines, conducted the analyses, and wrote the manuscript. E.-M. V. and M. W. contributed to the study design, the analysis and interpretation of the results, and the writing and revision of the manuscript. All authors read and approved the final manuscript.

\bibliography{refs}

\clearpage
\onecolumn
\newcounter{savedtable}\setcounter{savedtable}{\value{table}}
\newcounter{savedfigure}\setcounter{savedfigure}{\value{figure}}
\begin{appendices}
\setcounter{table}{\value{savedtable}}\setcounter{figure}{\value{savedfigure}}
\renewcommand{\thetable}{\arabic{table}}\renewcommand{\thefigure}{\arabic{figure}}

\section{News sample}\label{app:sample}

We draw the news test sample from the published release (version 2.0.0; 8{,}623 coded videos, of which the original coders classify 7{,}978 as news), using a class-quota sampler with seed 42. We target 200 videos per class of geographic scope (three classes), negative news, conflict, and scope. For each variable, we exclude videos with unclear codes from its sampling pools. At each step, the sampler selects the class furthest below its target, draws a candidate with probability proportional to its country's share of the news corpus, and counts that video toward every matching class still below target. We exclude 132 development videos that we drew earlier with the same sampler and smaller targets.

The sampler selects 612 test videos. We could not download nine because of age restrictions, blocked access, or failed writes, and we did not replace them, leaving 603. We reproduce the 612 selected videos exactly when we re-run the sampler with the archived settings. Because each video counts toward every matching class, the quotas over-represent minority classes rather than producing exact balance (Table~\ref{tab:newssample}). The sample's country mix (Switzerland 58\%, Austria 34\%, Germany 8\%) also differs from the news corpus (47\%, 38\%, 16\%). We use 599 videos in the regression analysis, excluding three without view counts and one with an unclear human code for geographic scope.

\begin{appendixtable}
\centering\footnotesize
\caption{Class shares in the news test sample and in all news videos of the published corpus. Binary constructs: share coded 1. Quota: construct used in drawing the sample.}
\label{tab:newssample}

\begin{tabular}{@{}llrr@{}}
\toprule
Construct & Class & Test sample (603) & News corpus (7{,}978)\\
\midrule
Geographic scope (quota) & 0 & 33\% & 14\%\\
 & 1 & 34\% & 56\%\\
 & 2 & 33\% & 30\%\\
Negative news (quota) & 1 & 33\% & 22\%\\
Conflict (quota) & 1 & 33\% & 14\%\\
Scope (quota) & 1 & 67\% & 81\%\\
Call to interact (TikTok) & 1 & 24\% & 26\%\\
Call to interact (off-platform) & 1 & 17\% & 17\%\\
Other people's voice & 1 & 25\% & 25\%\\
Visual set-up & 1 & 47\% & 50\%\\
 & 2 & 21\% & 20\%\\
 & 3 & 32\% & 30\%\\
Music & 1 & 69\% & 69\%\\
\bottomrule
\end{tabular}

\end{appendixtable}

\FloatBarrier
\needspace{0.30\textheight}
\section{Models and inference settings}\label{app:models}

\begin{appendixtable}
\centering\scriptsize
\caption{Model checkpoints. We load each model from the Hugging Face Hub at the listed revision and run it in bfloat16 with vLLM~0.30.0 on one NVIDIA A100 (80\,GB).}
\label{tab:models}
\fitwidth{
\begin{tabular}{@{}lll@{}}
\toprule
Model & Checkpoint (revision) & Architecture\\
\midrule
Gemma 4 12B & \texttt{google/gemma-4-12B-it} (707f0a3b8a) & dense\\
Gemma 4 E4B & \texttt{google/gemma-4-E4B-it} (ee0ef60236) & dense, edge-oriented\\
Qwen3-Omni & \texttt{Qwen/Qwen3-Omni-30B-A3B-Instruct} (26291f7938) & MoE, $\approx$3B active\\
Qwen2.5-Omni & \texttt{Qwen/Qwen2.5-Omni-7B} (ae9e169054) & dense\\
Nemotron-3 Nano Omni & \texttt{nvidia/Nemotron-3-Nano-Omni-30B-A3B-Reasoning-BF16} (e5e9932441) & hybrid Mamba--transformer MoE, $\approx$3B active\\
MiniCPM-o 4.5 & \texttt{openbmb/MiniCPM-o-4\_5} (503e754207) & dense\\
\bottomrule
\end{tabular}
}
\end{appendixtable}

\subsection*{Analysis history}
The frozen local plan specified the original text, frames + audio, combined, description, metadata, and human-transcript comparisons and identified the ablations, engagement regression, and re-run as exploratory. The switch to vLLM on the A100 and the additions of Qwen3-Omni, Qwen2.5-Omni, and MiniCPM-o followed inspection of interim laptop results; all reported primary predictions were generated anew on the A100. Native and full video, the HateMM target variables, the expanded audio diagnostics, and the integrated grid and downstream comparisons were subsequent extensions. The archived plan and deviation log document the original scope and these changes; they do not constitute a public preregistration of the final design.

\subsection*{Inference and preprocessing}
We use greedy decoding (temperature 0) for coding calls, with at most 120 output tokens and a JSON-schema grammar that lists each construct's legal integer codes. We disable reasoning modes through the chat template. For description calls in the describe-first route, we use greedy decoding with at most 450 tokens. We process requests in batches of 256 and disable prefix caching. We also disable the multimodal preprocessing cache except for capped full-video runs. We run Nemotron-3 Nano Omni with one image tile per frame.

We divide audio longer than 30 seconds into consecutive 30-second clips (at most four), matching the per-clip limit of the Gemma~4 audio encoder \citep{Google2026gemma4}. We transcribe speech with Whisper large-v3-turbo \citep{Radfordetal2023whisper} and automatic language detection. We extract on-screen text with Apple's Vision framework at one frame per second and remove near-duplicate lines. In the text prompts, we retain at most 6{,}000 characters of the automatic transcript, 3{,}000 characters of OCR text, and 6{,}000 characters of a generated description. For frames + audio, we sample the midpoints of sixteen equal segments of the first 120 seconds and scale the images to 448 pixels on the longest side.

For native video, we prepare a file containing the first 120 seconds at the same resolution and let each model's video loader sample sixteen frames. We supply audio as separate clips to Gemma~4, Qwen3-Omni, and MiniCPM-o; a vLLM error prevents Qwen3-Omni from using time-aligned audio. For Qwen2.5-Omni and Nemotron, we supply audio aligned with the video. For full video, we pass the original file and use each model's default sampling. We provide its audio as separate clips from the first 120 seconds for Gemma~4, Qwen3-Omni, and MiniCPM-o, and as time-aligned audio for Qwen2.5-Omni and Nemotron. We count processing failures as incorrect and omit full-video cells where failures exceed 5\%. All retained full-video cells use the original file. For capped runs, we use the same file, limit the video loader to 32 frames, and supply audio clips from the first 120 seconds. Capped-news tests cover 20 videos per Qwen model; capped HateMM tests cover all 400 videos for every model (Table~\ref{tab:feasibility}).

In the Gemma diagnostic tests, we add an audio-only route, follow the recommended media placement (frames before the instruction, audio after it), ask directly about music, and test transcription with the developer's prompt. We also repeat the tests with the Hugging Face transformers version installed in the vLLM image. We summarize the runtime configuration in Table~\ref{tab:runtime}. We did not log the PyTorch and transformers versions separately; we used the versions shipped with the public image \texttt{vllm/vllm-openai:v0.30.0}.

\begin{appendixtable}
\centering\scriptsize
\caption{Runtime configuration of all model calls and of the extraction steps.}
\label{tab:runtime}

\begin{tabular}{@{}lL{11.2cm}@{}}
\toprule
Component & Setting\\
\midrule
Inference engine & vLLM 0.30.0 (Docker image \texttt{vllm/vllm-openai:v0.30.0}; version confirmed in the engine logs); PyTorch and transformers as shipped in the image (versions not logged)\\
Hardware, precision & one NVIDIA A100-SXM4-80GB; bfloat16 weights and activations\\
Engine settings & seed 0; prefix caching off; multimodal preprocessing cache off (on only for the capped full-video runs); at most 32 images, 4 audio clips, 1 video per prompt; context 32{,}768 tokens (65{,}536 for long media inputs)\\
Prompting & each model's own chat template; reasoning disabled (\texttt{enable\_thinking=False}); news constructs jointly; HateMM targets jointly in a separate call from hate; media placed before the instruction (Gemma diagnosis: also after)\\
Decoding & greedy (temperature 0); at most 120 output tokens for codes, 450 for descriptions\\
Structured output & JSON schema per corpus restricting each construct to its legal codes (vLLM structured outputs, xgrammar backend)\\
Model-specific & Nemotron: remote code, one image tile per frame; MiniCPM-o: remote code; Qwen2.5-Omni and Nemotron: time-aligned audio in the native and full video routes; Qwen3-Omni: audio as separate clips there (vLLM 0.30 fails on its time-aligned audio); Gemma models and MiniCPM-o: audio as separate clips in every route\\
Media & frames 448\,px on the longest side, 16 evenly spaced over the first 120\,s (ablations: 4, 32; 896\,px); audio as up to four 30-s clips (16\,kHz, lossless FLAC) covering the first 120\,s; native video: first 120\,s at 448\,px (H.264, AAC)\\
Speech recognition & Whisper large-v3-turbo (\texttt{mlx-community/whisper-large-v3-turbo}, mlx-whisper), temperature 0, no conditioning on previous text, language detected automatically\\
OCR & Apple Vision text recognition (accurate level, language correction, confidence $\geq .3$) at one frame per second; near-duplicate lines removed\\
\bottomrule
\end{tabular}

\end{appendixtable}

\FloatBarrier
\needspace{0.75\textheight}
\section{Full results}\label{app:fullgrid}

\begin{appendixtable}
\centering\footnotesize
\caption{Macro-F1 by route, model, and construct group (mean of construct-level macro-F1). News, verbal: seven verbally defined constructs ($n=603$); anchors: visual set-up and music; hate: $n=400$; sarcasm: $n=650$. ``--'': route not defined for the dataset or not run.}
\label{tab:fullgrid}
\fitwidth{
\begin{tabular}{@{}llrrrrrrrrr@{}}
\toprule
Group & Model & \rotatebox{60}{Metadata} & \rotatebox{60}{Transcript} & \rotatebox{60}{Human transcr.} & \rotatebox{60}{Extracted text} & \rotatebox{60}{Frames + audio} & \rotatebox{60}{Native video} & \rotatebox{60}{Full video} & \rotatebox{60}{Text + video} & \rotatebox{60}{Describe first}\\
\midrule
News, 7 verbal & Gemma 4 12B & .69 & .79 & -- & .80 & .75 & .75 & .74 & .75 & .80\\
 & Gemma 4 E4B & .68 & .76 & -- & .76 & .73 & .73 & .73 & .74 & .76\\
 & Qwen3-Omni & .68 & .76 & -- & .76 & .71 & .70 & -- & .71 & .70\\
 & Qwen2.5-Omni & .49 & .52 & -- & .52 & .71 & .70 & -- & .73 & .49\\
 & Nemotron-3 Nano Omni & .55 & .63 & -- & .63 & .71 & .71 & .74 & .71 & .61\\
 & MiniCPM-o 4.5 & .61 & .70 & -- & .66 & .76 & .77 & .78 & .77 & .67\\
\addlinespace
News, 2 anchors & Gemma 4 12B & .39 & .47 & -- & .44 & .51 & .50 & .48 & .52 & .50\\
 & Gemma 4 E4B & .34 & .30 & -- & .35 & .50 & .50 & .50 & .51 & .46\\
 & Qwen3-Omni & .38 & .42 & -- & .39 & .67 & .62 & -- & .65 & .47\\
 & Qwen2.5-Omni & .30 & .28 & -- & .29 & .57 & .63 & -- & .59 & .47\\
 & Nemotron-3 Nano Omni & .26 & .27 & -- & .26 & .63 & .71 & .70 & .63 & .49\\
 & MiniCPM-o 4.5 & .45 & .32 & -- & .33 & .81 & .80 & .81 & .81 & .52\\
\addlinespace
Hate & Gemma 4 12B & -- & .82 & -- & .83 & .78 & .78 & .79 & .83 & .77\\
 & Gemma 4 E4B & -- & .80 & -- & .82 & .78 & .79 & .79 & .85 & .77\\
 & Qwen3-Omni & -- & .79 & -- & .80 & .74 & .77 & -- & .76 & .63\\
 & Qwen2.5-Omni & -- & .74 & -- & .76 & .77 & .76 & -- & .77 & .73\\
 & Nemotron-3 Nano Omni & -- & .75 & -- & .76 & .75 & .74 & .72 & .76 & .74\\
 & MiniCPM-o 4.5 & -- & .77 & -- & .78 & .77 & .80 & .80 & .76 & .78\\
\addlinespace
Target: Black & Gemma 4 12B & -- & .83 & -- & .83 & .77 & .77 & .77 & .82 & .73\\
 & Gemma 4 E4B & -- & .82 & -- & .83 & .71 & .75 & .73 & .83 & .77\\
 & Qwen3-Omni & -- & .80 & -- & .82 & .79 & .80 & -- & .80 & .63\\
 & Qwen2.5-Omni & -- & .81 & -- & .81 & .76 & .71 & -- & .80 & .68\\
 & Nemotron-3 Nano Omni & -- & .81 & -- & .83 & .80 & .78 & .78 & .82 & .80\\
 & MiniCPM-o 4.5 & -- & .77 & -- & .79 & .78 & .80 & .80 & .81 & .75\\
\addlinespace
Target: Jewish & Gemma 4 12B & -- & .76 & -- & .75 & .79 & .81 & .80 & .79 & .79\\
 & Gemma 4 E4B & -- & .79 & -- & .78 & .78 & .81 & .80 & .79 & .78\\
 & Qwen3-Omni & -- & .77 & -- & .77 & .78 & .77 & -- & .77 & .70\\
 & Qwen2.5-Omni & -- & .79 & -- & .78 & .73 & .72 & -- & .75 & .77\\
 & Nemotron-3 Nano Omni & -- & .77 & -- & .75 & .79 & .77 & .76 & .79 & .78\\
 & MiniCPM-o 4.5 & -- & .76 & -- & .73 & .80 & .82 & .80 & .80 & .78\\
\addlinespace
Sarcasm & Gemma 4 12B & .41 & .70 & .71 & .70 & .50 & .52 & .55 & .73 & .47\\
 & Gemma 4 E4B & .48 & .68 & .68 & .68 & .59 & .59 & .57 & .67 & .61\\
 & Qwen3-Omni & .52 & .47 & .47 & .47 & .58 & .60 & .63 & .51 & .59\\
 & Qwen2.5-Omni & .48 & .49 & .53 & .49 & .58 & .54 & .56 & .61 & .60\\
 & Nemotron-3 Nano Omni & .51 & .58 & .62 & .58 & .55 & .55 & .51 & .50 & .66\\
 & MiniCPM-o 4.5 & .51 & .36 & .37 & .36 & .40 & .47 & .47 & .36 & .55\\
\bottomrule
\end{tabular}
}
\end{appendixtable}

\begin{appendixtable}
\centering\footnotesize
\caption{Route choice across 78 model $\times$ construct combinations. Best: share in which the route has the highest macro-F1. Shortfall: macro-F1 below the best route of the same combination. Best or practically equivalent: the route is best, or the 90\% paired-bootstrap interval of its difference from the best lies within the margin. Worse: the interval lies entirely below $-.05$. Full video: 54 combinations; metadata only: 60; human transcript: sarcasm only. Descriptive: the comparator is the observed maximum of each combination.}
\label{tab:equivalence}
\fitwidth{
\begin{tabular}{@{}lrrrrrrrr@{}}
\toprule
 & & & \multicolumn{2}{c}{Shortfall} & \multicolumn{3}{c}{Best or practically equivalent} & Worse\\
\cmidrule(lr){4-5}\cmidrule(lr){6-8}\cmidrule(l){9-9}
Route & Cells & Best & Median & 10th--90th pct. & $\pm.03$ & $\pm.05$ & $\pm.10$ & $\pm.05$\\
\midrule
Metadata only & 60 & 7\% & .18 & .02--.35 & 8\% & 12\% & 23\% & 73\%\\
Transcript & 78 & 3\% & .04 & .00--.31 & 9\% & 32\% & 60\% & 35\%\\
Human transcript & 6 & 0\% & .06 & .01--.17 & 17\% & 33\% & 50\% & 50\%\\
Extracted text & 78 & 21\% & .04 & .00--.31 & 27\% & 37\% & 59\% & 35\%\\
Frames + audio & 78 & 18\% & .04 & .00--.12 & 24\% & 35\% & 73\% & 23\%\\
Native video & 78 & 19\% & .04 & .00--.11 & 27\% & 37\% & 73\% & 21\%\\
Full video & 54 & 13\% & .03 & .00--.11 & 30\% & 48\% & 72\% & 20\%\\
Text + video & 78 & 13\% & .02 & .00--.10 & 31\% & 55\% & 82\% & 15\%\\
Describe first & 78 & 13\% & .06 & .00--.25 & 13\% & 18\% & 53\% & 41\%\\
\bottomrule
\end{tabular}
}
\end{appendixtable}

\FloatBarrier
\needspace{0.50\textheight}
\section{Per-construct contrasts}\label{app:contrasts}

\begin{appendixfigure}
\centering
\includegraphics[width=\textwidth]{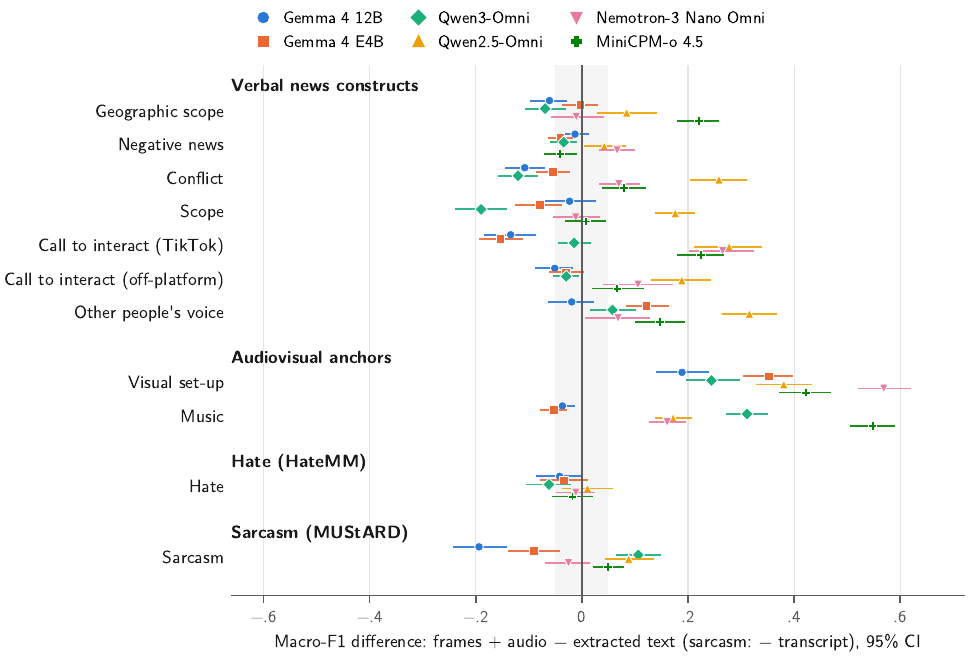}
\caption{Frames + audio minus extracted text (sarcasm: minus transcript) by construct and model, with 95\% item-bootstrap intervals. Shaded band: $\pm.05$.}
\label{fig:contrasts}
\end{appendixfigure}

\begin{appendixtable}
\centering\scriptsize
\caption{Native video minus extracted text (sarcasm: minus transcript), macro-F1, with 95\% item-bootstrap intervals. Bold: interval excludes zero.}
\label{tab:contrastsnative}
\fitwidth{
\begin{tabular}{@{}lcccccc@{}}
\toprule
Construct & Gemma 12B & Gemma E4B & Qwen3-Omni & Qwen2.5-Omni & Nemotron & MiniCPM-o\\
\midrule
Geographic scope & \textbf{$-$.07} & +.01 & \textbf{$-$.06} & +.03 & +.04 & \textbf{+.21}\\
 & {\scriptsize[$-$.10, $-$.04]} & {\scriptsize[$-$.02, +.04]} & {\scriptsize[$-$.09, $-$.02]} & {\scriptsize[$-$.03, +.09]} & {\scriptsize[$-$.01, +.08]} & {\scriptsize[+.17, +.25]}\\[2pt]
Negative news & $-$.01 & \textbf{$-$.03} & $-$.01 & \textbf{+.04} & \textbf{+.05} & \textbf{$-$.04}\\
 & {\scriptsize[$-$.03, +.02]} & {\scriptsize[$-$.05, $-$.01]} & {\scriptsize[$-$.03, +.02]} & {\scriptsize[+.00, +.08]} & {\scriptsize[+.02, +.08]} & {\scriptsize[$-$.07, $-$.01]}\\[2pt]
Conflict & \textbf{$-$.08} & \textbf{$-$.04} & \textbf{$-$.09} & \textbf{+.28} & +.02 & \textbf{+.08}\\
 & {\scriptsize[$-$.11, $-$.04]} & {\scriptsize[$-$.06, $-$.01]} & {\scriptsize[$-$.13, $-$.06]} & {\scriptsize[+.23, +.32]} & {\scriptsize[$-$.02, +.06]} & {\scriptsize[+.04, +.12]}\\[2pt]
Scope & \textbf{$-$.08} & \textbf{$-$.07} & \textbf{$-$.18} & \textbf{+.14} & +.00 & +.02\\
 & {\scriptsize[$-$.13, $-$.03]} & {\scriptsize[$-$.11, $-$.03]} & {\scriptsize[$-$.23, $-$.14]} & {\scriptsize[+.10, +.17]} & {\scriptsize[$-$.04, +.05]} & {\scriptsize[$-$.02, +.05]}\\[2pt]
Call to interact (TikTok) & $-$.03 & \textbf{$-$.12} & $-$.01 & \textbf{+.38} & \textbf{+.27} & \textbf{+.22}\\
 & {\scriptsize[$-$.07, +.00]} & {\scriptsize[$-$.16, $-$.08]} & {\scriptsize[$-$.04, +.02]} & {\scriptsize[+.33, +.44]} & {\scriptsize[+.22, +.32]} & {\scriptsize[+.17, +.26]}\\[2pt]
Call to interact (off-platform) & \textbf{$-$.06} & \textbf{$-$.04} & $-$.02 & \textbf{+.19} & $-$.03 & \textbf{+.12}\\
 & {\scriptsize[$-$.09, $-$.03]} & {\scriptsize[$-$.07, $-$.01]} & {\scriptsize[$-$.04, +.01]} & {\scriptsize[+.13, +.24]} & {\scriptsize[$-$.11, +.04]} & {\scriptsize[+.07, +.18]}\\[2pt]
Other people's voice & \textbf{$-$.08} & \textbf{+.05} & \textbf{$-$.05} & \textbf{+.22} & \textbf{+.20} & \textbf{+.18}\\
 & {\scriptsize[$-$.12, $-$.03]} & {\scriptsize[+.01, +.08]} & {\scriptsize[$-$.10, $-$.01]} & {\scriptsize[+.18, +.26]} & {\scriptsize[+.15, +.25]} & {\scriptsize[+.13, +.22]}\\[2pt]
Visual set-up & \textbf{+.16} & \textbf{+.34} & \textbf{+.19} & \textbf{+.43} & \textbf{+.58} & \textbf{+.40}\\
 & {\scriptsize[+.11, +.21]} & {\scriptsize[+.30, +.39]} & {\scriptsize[+.14, +.24]} & {\scriptsize[+.38, +.48]} & {\scriptsize[+.54, +.63]} & {\scriptsize[+.35, +.45]}\\[2pt]
Music & $-$.02 & \textbf{$-$.04} & \textbf{+.26} & \textbf{+.24} & \textbf{+.31} & \textbf{+.56}\\
 & {\scriptsize[$-$.05, +.00]} & {\scriptsize[$-$.07, $-$.02]} & {\scriptsize[+.23, +.30]} & {\scriptsize[+.21, +.28]} & {\scriptsize[+.27, +.35]} & {\scriptsize[+.51, +.60]}\\[2pt]
Hate & $-$.04 & $-$.03 & $-$.03 & $-$.00 & $-$.03 & +.01\\
 & {\scriptsize[$-$.09, +.00]} & {\scriptsize[$-$.07, +.02]} & {\scriptsize[$-$.06, +.01]} & {\scriptsize[$-$.05, +.05]} & {\scriptsize[$-$.06, +.01]} & {\scriptsize[$-$.02, +.05]}\\[2pt]
Targets Black people & \textbf{$-$.06} & \textbf{$-$.08} & $-$.02 & \textbf{$-$.10} & \textbf{$-$.05} & +.01\\
 & {\scriptsize[$-$.10, $-$.02]} & {\scriptsize[$-$.12, $-$.04]} & {\scriptsize[$-$.06, +.02]} & {\scriptsize[$-$.15, $-$.05]} & {\scriptsize[$-$.09, $-$.01]} & {\scriptsize[$-$.02, +.06]}\\[2pt]
Targets Jewish people & \textbf{+.05} & +.03 & $-$.00 & \textbf{$-$.06} & +.02 & \textbf{+.08}\\
 & {\scriptsize[+.00, +.10]} & {\scriptsize[$-$.02, +.09]} & {\scriptsize[$-$.04, +.04]} & {\scriptsize[$-$.11, $-$.01]} & {\scriptsize[$-$.03, +.07]} & {\scriptsize[+.04, +.12]}\\[2pt]
Sarcasm & \textbf{$-$.18} & \textbf{$-$.09} & \textbf{+.12} & \textbf{+.04} & $-$.03 & \textbf{+.11}\\
 & {\scriptsize[$-$.23, $-$.13]} & {\scriptsize[$-$.14, $-$.05]} & {\scriptsize[+.08, +.17]} & {\scriptsize[+.00, +.09]} & {\scriptsize[$-$.07, +.01]} & {\scriptsize[+.08, +.14]}\\[2pt]
\bottomrule
\end{tabular}
}
\end{appendixtable}

\begin{appendixtable}
\centering\scriptsize
\caption{Frames + audio minus extracted text (sarcasm: minus transcript), macro-F1, with 95\% item-bootstrap intervals. Bold: interval excludes zero. These are the values plotted in Figure~\ref{fig:contrasts}.}
\label{tab:contrastsfa}
\fitwidth{
\begin{tabular}{@{}lcccccc@{}}
\toprule
Construct & Gemma 12B & Gemma E4B & Qwen3-Omni & Qwen2.5-Omni & Nemotron & MiniCPM-o\\
\midrule
Geographic scope & \textbf{$-$.06} & $-$.00 & \textbf{$-$.07} & \textbf{+.08} & $-$.01 & \textbf{+.22}\\
 & {\scriptsize[$-$.10, $-$.03]} & {\scriptsize[$-$.04, +.03]} & {\scriptsize[$-$.11, $-$.03]} & {\scriptsize[+.03, +.14]} & {\scriptsize[$-$.06, +.04]} & {\scriptsize[+.18, +.26]}\\[2pt]
Negative news & $-$.01 & \textbf{$-$.04} & \textbf{$-$.03} & \textbf{+.04} & \textbf{+.07} & \textbf{$-$.04}\\
 & {\scriptsize[$-$.04, +.01]} & {\scriptsize[$-$.06, $-$.02]} & {\scriptsize[$-$.06, $-$.01]} & {\scriptsize[+.01, +.08]} & {\scriptsize[+.03, +.10]} & {\scriptsize[$-$.07, $-$.01]}\\[2pt]
Conflict & \textbf{$-$.11} & \textbf{$-$.05} & \textbf{$-$.12} & \textbf{+.26} & \textbf{+.07} & \textbf{+.08}\\
 & {\scriptsize[$-$.14, $-$.07]} & {\scriptsize[$-$.08, $-$.02]} & {\scriptsize[$-$.16, $-$.08]} & {\scriptsize[+.21, +.31]} & {\scriptsize[+.03, +.11]} & {\scriptsize[+.04, +.12]}\\[2pt]
Scope & $-$.02 & \textbf{$-$.08} & \textbf{$-$.19} & \textbf{+.18} & $-$.01 & +.01\\
 & {\scriptsize[$-$.07, +.02]} & {\scriptsize[$-$.12, $-$.04]} & {\scriptsize[$-$.24, $-$.14]} & {\scriptsize[+.14, +.21]} & {\scriptsize[$-$.05, +.03]} & {\scriptsize[$-$.03, +.04]}\\[2pt]
Call to interact (TikTok) & \textbf{$-$.13} & \textbf{$-$.15} & $-$.01 & \textbf{+.28} & \textbf{+.27} & \textbf{+.22}\\
 & {\scriptsize[$-$.18, $-$.09]} & {\scriptsize[$-$.19, $-$.11]} & {\scriptsize[$-$.04, +.02]} & {\scriptsize[+.21, +.34]} & {\scriptsize[+.20, +.32]} & {\scriptsize[+.18, +.27]}\\[2pt]
Call to interact (off-platform) & \textbf{$-$.05} & $-$.03 & \textbf{$-$.03} & \textbf{+.19} & \textbf{+.11} & \textbf{+.07}\\
 & {\scriptsize[$-$.09, $-$.02]} & {\scriptsize[$-$.06, +.00]} & {\scriptsize[$-$.05, $-$.01]} & {\scriptsize[+.13, +.24]} & {\scriptsize[+.04, +.17]} & {\scriptsize[+.02, +.12]}\\[2pt]
Other people's voice & $-$.02 & \textbf{+.12} & \textbf{+.06} & \textbf{+.32} & \textbf{+.07} & \textbf{+.15}\\
 & {\scriptsize[$-$.06, +.02]} & {\scriptsize[+.09, +.16]} & {\scriptsize[+.02, +.10]} & {\scriptsize[+.27, +.37]} & {\scriptsize[+.01, +.13]} & {\scriptsize[+.10, +.19]}\\[2pt]
Visual set-up & \textbf{+.19} & \textbf{+.35} & \textbf{+.24} & \textbf{+.38} & \textbf{+.57} & \textbf{+.42}\\
 & {\scriptsize[+.14, +.24]} & {\scriptsize[+.31, +.40]} & {\scriptsize[+.20, +.30]} & {\scriptsize[+.33, +.43]} & {\scriptsize[+.52, +.62]} & {\scriptsize[+.37, +.47]}\\[2pt]
Music & \textbf{$-$.04} & \textbf{$-$.05} & \textbf{+.31} & \textbf{+.17} & \textbf{+.16} & \textbf{+.55}\\
 & {\scriptsize[$-$.06, $-$.01]} & {\scriptsize[$-$.08, $-$.03]} & {\scriptsize[+.27, +.35]} & {\scriptsize[+.14, +.21]} & {\scriptsize[+.13, +.19]} & {\scriptsize[+.51, +.59]}\\[2pt]
Hate & $-$.04 & $-$.03 & \textbf{$-$.06} & +.01 & $-$.01 & $-$.02\\
 & {\scriptsize[$-$.09, +.00]} & {\scriptsize[$-$.08, +.01]} & {\scriptsize[$-$.10, $-$.02]} & {\scriptsize[$-$.04, +.06]} & {\scriptsize[$-$.05, +.02]} & {\scriptsize[$-$.05, +.02]}\\[2pt]
Targets Black people & \textbf{$-$.06} & \textbf{$-$.12} & $-$.03 & \textbf{$-$.06} & $-$.03 & $-$.01\\
 & {\scriptsize[$-$.11, $-$.02]} & {\scriptsize[$-$.16, $-$.08]} & {\scriptsize[$-$.07, +.01]} & {\scriptsize[$-$.11, $-$.00]} & {\scriptsize[$-$.07, +.01]} & {\scriptsize[$-$.05, +.04]}\\[2pt]
Targets Jewish people & +.04 & $-$.00 & +.01 & \textbf{$-$.05} & +.04 & \textbf{+.06}\\
 & {\scriptsize[$-$.02, +.09]} & {\scriptsize[$-$.06, +.06]} & {\scriptsize[$-$.03, +.05]} & {\scriptsize[$-$.10, $-$.00]} & {\scriptsize[$-$.01, +.08]} & {\scriptsize[+.02, +.11]}\\[2pt]
Sarcasm & \textbf{$-$.19} & \textbf{$-$.09} & \textbf{+.11} & \textbf{+.09} & $-$.03 & \textbf{+.05}\\
 & {\scriptsize[$-$.24, $-$.14]} & {\scriptsize[$-$.14, $-$.04]} & {\scriptsize[+.07, +.15]} & {\scriptsize[+.05, +.13]} & {\scriptsize[$-$.07, +.01]} & {\scriptsize[+.02, +.08]}\\[2pt]
\bottomrule
\end{tabular}
}
\end{appendixtable}

\begin{appendixtable}
\centering\footnotesize
\caption{Construct dependence on the same items: difference-in-differences of the video-minus-extracted-text contrast between two constructs coded on the same items, macro-F1, with 95\% paired item-bootstrap intervals (1{,}000 resamples). News: music minus the mean of the seven verbal constructs (603 videos). HateMM: targets Jewish people minus targets Black people (400 videos). Bold: interval excludes zero.}
\label{tab:did}
\fitwidth{
\begin{tabular}{@{}lllll@{}}
\toprule
 & \multicolumn{2}{c}{News: music $-$ verbal mean} & \multicolumn{2}{c}{HateMM: Jewish $-$ Black target}\\
\cmidrule(lr){2-3}\cmidrule(l){4-5}
Model & Frames + audio & Native video & Frames + audio & Native video\\
\midrule
Gemma 4 12B & .02 [$-$.00, .05] & \textbf{.04 [.01, .07]} & \textbf{.10 [.03, .17]} & \textbf{.11 [.05, .18]}\\
Gemma 4 E4B & $-$.02 [$-$.04, .01] & $-$.01 [$-$.04, .02] & \textbf{.12 [.05, .20]} & \textbf{.11 [.04, .18]}\\
Qwen3-Omni & \textbf{.37 [.33, .41]} & \textbf{.32 [.28, .36]} & .04 [$-$.02, .10] & .02 [$-$.04, .08]\\
Qwen2.5-Omni & $-$.02 [$-$.06, .02] & \textbf{.06 [.02, .10]} & .00 [$-$.07, .07] & .04 [$-$.03, .12]\\
Nemotron-3 Nano Omni & \textbf{.08 [.05, .12]} & \textbf{.23 [.19, .27]} & \textbf{.07 [.00, .13]} & .07 [$-$.00, .14]\\
MiniCPM-o 4.5 & \textbf{.45 [.41, .49]} & \textbf{.45 [.40, .49]} & \textbf{.07 [.01, .14]} & \textbf{.07 [.01, .13]}\\
\bottomrule
\end{tabular}
}
\end{appendixtable}

Pooled over the seven verbal news constructs, native video and frames + audio differ by at most .012 in any model. Individual constructs move more: the two video representations differ by up to $\pm.15$ for other people's voice, music, and calls to interact off-platform (\texttt{results/construct\_contrasts.csv}).

\FloatBarrier
\needspace{0.25\textheight}
\section{Video input, resolution, and frame budget}\label{app:videoinput}

\begin{appendixtable}
\centering\footnotesize
\caption{Macro-F1 by form of video input. 896\,px: difference to 448-pixel frames on a random subsample of 150 news videos. $^a$Difference, not level. $^b$The original file does not fit the model for most videos (context window or GPU memory).}
\label{tab:videoinput}
\fitwidth{
\begin{tabular}{@{}lccccc@{}}
\toprule
 & \multicolumn{4}{c}{News, 7 verbal constructs} & \\ \cmidrule(lr){2-5}
Model & Frames + audio & Native video & Full video & 896\,px frames$^a$ & Sarcasm: native / full\\
\midrule
Gemma 4 12B & .75 & .75 & .74 & +.00 & .52 / .55\\
Gemma 4 E4B & .73 & .73 & .73 & $-$.00 & .59 / .57\\
Qwen3-Omni & .71 & .70 & n/a$^b$ & +.00 & .60 / .63\\
Qwen2.5-Omni & .71 & .70 & n/a$^b$ & $-$.00 & .54 / .56\\
Nemotron-3 Nano Omni & .71 & .71 & .74 & +.02 & .55 / .51\\
MiniCPM-o 4.5 & .76 & .77 & .78 & $-$.00 & .47 / .47\\
\bottomrule
\end{tabular}
}
\end{appendixtable}

\begin{appendixtable}
\centering\footnotesize
\caption{Audio and frame ablations on 150 news videos (macro-F1).}
\label{tab:ablation}

\begin{tabular}{@{}lcccc|cc@{}}
\toprule
 & \multicolumn{4}{c|}{Seven verbal news constructs} & \multicolumn{2}{c}{Music}\\
Model & 16 frames + audio & no audio & 4 frames & 32 frames & with audio & no audio\\
\midrule
Gemma 4 12B & .74 & .74 & .71 & .74 & .22 & .23\\
Gemma 4 E4B & .73 & .74 & .71 & .72 & .22 & .22\\
Qwen3-Omni & .71 & .70 & .71 & .70 & .59 & .28\\
Qwen2.5-Omni & .70 & .69 & .66 & .71 & .38 & .22\\
Nemotron-3 Nano Omni & .71 & .64 & .68 & .71 & .39 & .23\\
MiniCPM-o 4.5 & .76 & .71 & .72 & .75 & .84 & .38\\
\bottomrule
\end{tabular}

\end{appendixtable}

\begin{appendixtable}
\centering\footnotesize
\caption{Full video: successfully processed / attempted items for the original file (left) and capped version (right). Capped news runs are 20-video pilots, not full-sample evaluations. Hate counts refer to the hate-classification call; Qwen3-Omni processed no target-group calls with original or capped full video, while other models had the same completion counts for hate and targets. ``--'': not run.}
\label{tab:feasibility}

\begin{tabular}{@{}lccc|ccc@{}}
\toprule
 & \multicolumn{3}{c|}{Original file processed} & \multicolumn{3}{c}{Capped version processed}\\
Model & News & Hate & Sarcasm & News & Hate & Sarcasm\\
\midrule
Gemma 4 12B & 603/603 & 400/400 & 650/650 & -- & 400/400 & --\\
Gemma 4 E4B & 603/603 & 400/400 & 650/650 & -- & 400/400 & --\\
Qwen3-Omni & 0/603 & 1/400 & 650/650 & 0/20 & 1/400 & --\\
Qwen2.5-Omni & 224/603 & 145/400 & 650/650 & 12/20 & 147/400 & --\\
Nemotron-3 Nano Omni & 598/603 & 382/400 & 650/650 & -- & 400/400 & --\\
MiniCPM-o 4.5 & 603/603 & 400/400 & 650/650 & -- & 400/400 & --\\
\bottomrule
\end{tabular}

\end{appendixtable}

\begin{appendixtable}
\centering\footnotesize
\caption{Audio ablation on all datasets: macro-F1 of frames + audio $\rightarrow$ the same frames without audio.}
\label{tab:ablationall}
\fitwidth{
\begin{tabular}{@{}lcccc@{}}
\toprule
Model & News, verbal & Music & Hate & Sarcasm\\
\midrule
Gemma 4 12B & .74 $\rightarrow$ .74 & .22 $\rightarrow$ .23 & .78 $\rightarrow$ .64 & .50 $\rightarrow$ .43\\
Gemma 4 E4B & .73 $\rightarrow$ .74 & .22 $\rightarrow$ .22 & .78 $\rightarrow$ .57 & .59 $\rightarrow$ .43\\
Qwen3-Omni & .71 $\rightarrow$ .70 & .59 $\rightarrow$ .28 & .74 $\rightarrow$ .64 & .58 $\rightarrow$ .54\\
Qwen2.5-Omni & .70 $\rightarrow$ .69 & .38 $\rightarrow$ .22 & .77 $\rightarrow$ .61 & .58 $\rightarrow$ .51\\
Nemotron-3 Nano Omni & .71 $\rightarrow$ .64 & .39 $\rightarrow$ .23 & .75 $\rightarrow$ .61 & .55 $\rightarrow$ .55\\
MiniCPM-o 4.5 & .76 $\rightarrow$ .71 & .84 $\rightarrow$ .38 & .77 $\rightarrow$ .66 & .40 $\rightarrow$ .52\\
\bottomrule
\end{tabular}
}
\end{appendixtable}

\FloatBarrier
\needspace{0.25\textheight}
\section{Describe first}\label{app:describe}

\begin{appendixtable}
\centering\footnotesize
\caption{Describe first minus frames + audio (macro-F1, 95\% item-bootstrap interval). Relative to extracted text, describe first is practically equivalent on the verbal news constructs for five of six models.}
\label{tab:describe}

\begin{tabular}{@{}lcccc@{}}
\toprule
Model & News, verbal & News, anchors & Hate & Sarcasm\\
\midrule
Gemma 4 12B & +.05 {\scriptsize[+.04, +.07]} & $-$.02 {\scriptsize[$-$.04, +.01]} & $-$.01 {\scriptsize[$-$.06, +.03]} & $-$.03 {\scriptsize[$-$.07, +.01]}\\
Gemma 4 E4B & +.03 {\scriptsize[+.02, +.05]} & $-$.04 {\scriptsize[$-$.07, $-$.02]} & $-$.01 {\scriptsize[$-$.05, +.03]} & +.03 {\scriptsize[$-$.01, +.07]}\\
Qwen3-Omni & $-$.01 {\scriptsize[$-$.02, +.01]} & $-$.20 {\scriptsize[$-$.24, $-$.17]} & $-$.10 {\scriptsize[$-$.16, $-$.04]} & +.01 {\scriptsize[$-$.03, +.06]}\\
Qwen2.5-Omni & $-$.21 {\scriptsize[$-$.23, $-$.20]} & $-$.10 {\scriptsize[$-$.12, $-$.07]} & $-$.04 {\scriptsize[$-$.08, +.01]} & +.02 {\scriptsize[$-$.03, +.07]}\\
Nemotron-3 Nano Omni & $-$.10 {\scriptsize[$-$.12, $-$.08]} & $-$.14 {\scriptsize[$-$.17, $-$.11]} & $-$.01 {\scriptsize[$-$.05, +.03]} & +.10 {\scriptsize[+.06, +.15]}\\
MiniCPM-o 4.5 & $-$.09 {\scriptsize[$-$.11, $-$.08]} & $-$.29 {\scriptsize[$-$.32, $-$.26]} & +.01 {\scriptsize[$-$.03, +.06]} & +.14 {\scriptsize[+.11, +.18]}\\
\bottomrule
\end{tabular}

\end{appendixtable}

\FloatBarrier
\needspace{0.75\textheight}
\section{Integration}\label{app:integration}

\begin{appendixtable}
\centering\scriptsize
\caption{Integration across datasets: share of item $\times$ construct decisions correct only from extracted text (sarcasm: transcript), only from frames + audio, or wrong in both, and macro-F1 of each route, of both in one call, and of the oracle.}
\label{tab:integrationall}
\fitwidth{
\begin{tabular}{@{}llccc|cccc@{}}
\toprule
 & & \multicolumn{3}{c|}{\% of decisions correct only by / wrong by} & \multicolumn{4}{c}{Macro-F1}\\
Group & Model & Text only & Video only & Both wrong & Text & Video & Text + video & Oracle\\
\midrule
News, verbal & Gemma 4 12B & 10 & 5 & 11 & .80 & .75 & .75 & .87\\
 & Gemma 4 E4B & 9 & 6 & 15 & .76 & .73 & .74 & .83\\
 & Qwen3-Omni & 9 & 5 & 15 & .76 & .71 & .71 & .82\\
 & Qwen2.5-Omni & 9 & 26 & 13 & .52 & .71 & .73 & .82\\
 & Nemotron-3 Nano Omni & 10 & 20 & 12 & .63 & .71 & .71 & .85\\
 & MiniCPM-o 4.5 & 8 & 16 & 14 & .66 & .76 & .77 & .84\\
\addlinespace
News, anchors & Gemma 4 12B & 6 & 13 & 38 & .44 & .51 & .52 & .58\\
 & Gemma 4 E4B & 4 & 18 & 40 & .35 & .50 & .51 & .56\\
 & Qwen3-Omni & 4 & 30 & 26 & .39 & .67 & .65 & .72\\
 & Qwen2.5-Omni & 4 & 32 & 35 & .29 & .57 & .59 & .62\\
 & Nemotron-3 Nano Omni & 4 & 35 & 32 & .26 & .63 & .63 & .67\\
 & MiniCPM-o 4.5 & 5 & 47 & 12 & .33 & .81 & .81 & .87\\
\addlinespace
Hate & Gemma 4 12B & 12 & 8 & 10 & .83 & .78 & .83 & .90\\
 & Gemma 4 E4B & 11 & 8 & 10 & .82 & .78 & .85 & .90\\
 & Qwen3-Omni & 10 & 4 & 15 & .80 & .74 & .76 & .84\\
 & Qwen2.5-Omni & 10 & 11 & 13 & .76 & .77 & .77 & .87\\
 & Nemotron-3 Nano Omni & 6 & 6 & 18 & .76 & .75 & .76 & .82\\
 & MiniCPM-o 4.5 & 7 & 6 & 16 & .78 & .77 & .76 & .84\\
\addlinespace
Target: Black people & Gemma 4 12B & 14 & 7 & 10 & .83 & .77 & .82 & .90\\
 & Gemma 4 E4B & 16 & 4 & 13 & .83 & .71 & .83 & .87\\
 & Qwen3-Omni & 9 & 7 & 11 & .82 & .79 & .80 & .89\\
 & Qwen2.5-Omni & 15 & 10 & 9 & .81 & .76 & .80 & .91\\
 & Nemotron-3 Nano Omni & 11 & 8 & 9 & .83 & .80 & .82 & .91\\
 & MiniCPM-o 4.5 & 9 & 8 & 12 & .79 & .78 & .81 & .87\\
\addlinespace
Target: Jewish people & Gemma 4 12B & 3 & 8 & 7 & .75 & .79 & .79 & .86\\
 & Gemma 4 E4B & 4 & 6 & 6 & .78 & .78 & .79 & .87\\
 & Qwen3-Omni & 3 & 4 & 10 & .77 & .78 & .77 & .82\\
 & Qwen2.5-Omni & 9 & 4 & 8 & .78 & .73 & .75 & .85\\
 & Nemotron-3 Nano Omni & 3 & 6 & 8 & .75 & .79 & .79 & .85\\
 & MiniCPM-o 4.5 & 2 & 8 & 8 & .73 & .80 & .80 & .84\\
\addlinespace
Sarcasm & Gemma 4 12B & 24 & 10 & 19 & .70 & .50 & .73 & .80\\
 & Gemma 4 E4B & 24 & 16 & 16 & .68 & .59 & .67 & .84\\
 & Qwen3-Omni & 4 & 10 & 33 & .47 & .58 & .51 & .63\\
 & Qwen2.5-Omni & 16 & 18 & 25 & .49 & .58 & .61 & .74\\
 & Nemotron-3 Nano Omni & 10 & 8 & 30 & .58 & .55 & .50 & .67\\
 & MiniCPM-o 4.5 & 1 & 3 & 46 & .36 & .40 & .36 & .42\\
\bottomrule
\end{tabular}
}
\end{appendixtable}

\begin{appendixtable}
\centering\footnotesize
\caption{Decision decomposition and oracle for extracted text versus native video (seven verbal news constructs). The main text uses frames + audio because text + video combines extracted text with that representation; the pattern is the same.}
\label{tab:integrationnative}
\fitwidth{
\begin{tabular}{@{}lcccc|ccc@{}}
\toprule
 & \multicolumn{4}{c|}{\% of decisions} & \multicolumn{3}{c}{Macro-F1}\\
Model & Both correct & Text only & Video only & Both wrong & Extracted text & Native video & Oracle\\
\midrule
Gemma 4 12B & 75.3 & 8.3 & 4.4 & 12.0 & .80 & .75 & .86\\
Gemma 4 E4B & 70.4 & 7.8 & 5.4 & 16.4 & .76 & .73 & .82\\
Qwen3-Omni & 70.8 & 9.3 & 4.8 & 15.1 & .76 & .70 & .82\\
Qwen2.5-Omni & 52.4 & 8.4 & 22.6 & 16.6 & .52 & .70 & .79\\
Nemotron-3 Nano Omni & 57.7 & 9.9 & 19.7 & 12.7 & .63 & .71 & .85\\
MiniCPM-o 4.5 & 63.2 & 6.8 & 15.6 & 14.5 & .66 & .77 & .84\\
\bottomrule
\end{tabular}
}
\end{appendixtable}

\FloatBarrier
\needspace{0.32\textheight}
\section{Prevalence}\label{app:prevalence}

\begin{appendixtable}
\centering\footnotesize
\caption{Prevalence error by construct and route: median across available models of estimated minus reference-coded sample prevalence (percentage points). Full video uses four models for news and HateMM and six for sarcasm; other defined routes use six.}
\label{tab:prevalence}
\fitwidth{
\begin{tabular}{@{}lcrrrrrrrr@{}}
\toprule
Construct & Reference prevalence & Metadata & Transcript & Extr.\ text & Fr.\,+\,audio & Native & Full & Text\,+\,video & Describe\\
\midrule
Negative news & 33\% & +12 & +20 & +23 & +27 & +26 & +27 & +26 & +17\\
Conflict & 33\% & $-$7 & $-$1 & +1 & +16 & +16 & +16 & +17 & $-$1\\
Scope & 67\% & $-$19 & $-$3 & +3 & $-$3 & +12 & +9 & +4 & $-$7\\
Call to interact (TikTok) & 24\% & +12 & +17 & +28 & $-$2 & +3 & +4 & +0 & +8\\
Call to interact (off-platform) & 17\% & $-$10 & $-$4 & $-$0 & $-$4 & $-$3 & $-$2 & $-$4 & $-$9\\
Other people's voice & 25\% & +7 & +4 & +19 & +15 & +26 & +7 & +11 & +20\\
Music & 69\% & $-$58 & $-$66 & $-$66 & $-$55 & $-$43 & $-$54 & $-$55 & $-$60\\
Hate & 50\% & $-$$-$ & +8 & +14 & +11 & +9 & +8 & +18 & $-$6\\
Targets Black people & 52\% & $-$$-$ & $-$4 & +1 & +6 & $-$0 & +0 & +11 & $-$15\\
Targets Jewish people & 12\% & $-$$-$ & +7 & +10 & +8 & +8 & +6 & +11 & +2\\
Sarcasm & 50\% & $-$1 & +35 & +35 & +21 & +31 & +27 & +32 & +5\\
\bottomrule
\end{tabular}
}
\end{appendixtable}

\begin{appendixtable}
\centering\footnotesize
\caption{Absolute prevalence error under four analysis sets. A: all routes. B: without metadata only and counting the identical sarcasm transcript/extracted-text route once. C: deployable routes (transcript, extracted text, frames + audio, native video, full video, text + video, describe first; the sarcasm transcript counted once). D: for each model and construct, the deployable route with the highest macro-F1. Error relative to the reference-coded prevalence of the same sample; the samples are balanced or quota samples.}
\label{tab:prevrobust}
\fitwidth{
\begin{tabular}{@{}llrrrrr@{}}
\toprule
Analysis set & Group & Estimates & Median abs.\ error (pp) & $>5$ pp & $>10$ pp & $>20$ pp\\
\midrule
all routes & All & 496 & 12.9 & 79\% & 59\% & 32\%\\
 & Hate and targets & 120 & 8.7 & 78\% & 42\% & 3\%\\
 & News, music & 46 & 64.3 & 91\% & 89\% & 89\%\\
 & News, verbal & 276 & 12.4 & 75\% & 57\% & 29\%\\
 & Sarcasm & 54 & 31.5 & 94\% & 81\% & 65\%\\
\addlinespace
without metadata only & All & 442 & 12.9 & 78\% & 59\% & 32\%\\
 & Hate and targets & 120 & 8.7 & 78\% & 42\% & 3\%\\
 & News, music & 40 & 65.0 & 90\% & 90\% & 90\%\\
 & News, verbal & 240 & 12.8 & 73\% & 58\% & 30\%\\
 & Sarcasm & 42 & 32.5 & 95\% & 86\% & 69\%\\
\addlinespace
deployable routes & All & 436 & 12.7 & 78\% & 59\% & 31\%\\
 & Hate and targets & 120 & 8.7 & 78\% & 42\% & 3\%\\
 & News, music & 40 & 65.0 & 90\% & 90\% & 90\%\\
 & News, verbal & 240 & 12.8 & 73\% & 58\% & 30\%\\
 & Sarcasm & 36 & 32.5 & 97\% & 86\% & 69\%\\
\addlinespace
highest-F1 deployable route & All & 66 & 9.0 & 65\% & 45\% & 21\%\\
 & Hate and targets & 18 & 5.3 & 56\% & 28\% & 0\%\\
 & News, music & 6 & 40.0 & 83\% & 83\% & 83\%\\
 & News, verbal & 36 & 9.0 & 61\% & 44\% & 17\%\\
 & Sarcasm & 6 & 18.7 & 100\% & 67\% & 50\%\\
\bottomrule
\end{tabular}
}
\end{appendixtable}

\FloatBarrier
\needspace{0.27\textheight}
\section{Agreement with human coders}\label{app:alpha}

\begin{appendixtable}
\centering\footnotesize
\caption{Krippendorff's $\alpha$ for the news constructs: published human pretest (two trained coders, 51 videos), best automated route, range across models for frames + audio, and agreement of the six models with each other on frames + audio.}
\label{tab:alpha}
\fitwidth{
\begin{tabular}{@{}lcccc@{}}
\toprule
Construct & Human pretest $\alpha$ & Best route $\alpha$ & Range, frames + audio & Six models as coders\\
\midrule
Geographic scope & .97 & .70 & .43--.70 & .61\\
Negative news & .91 & .60 & .31--.60 & .70\\
Conflict & .84 & .76 & .48--.67 & .63\\
Scope & 1.00 & .52 & $-$.11--.41 & .27\\
Call to interact (TikTok) & .91 & .71 & .38--.57 & .37\\
Call to interact (off-platform) & .82 & .87 & .28--.81 & .64\\
Other people's voice & .95 & .71 & .24--.62 & .37\\
\bottomrule
\end{tabular}
}
\end{appendixtable}

\FloatBarrier
\needspace{0.25\textheight}
\section{Re-run consistency}\label{app:rerun}

\begin{appendixtable}
\centering\footnotesize
\caption{Re-run of 50 news videos on six routes (metadata only, transcript, extracted text, frames + audio, text + video, describe first) with identical settings on the same GPU.}
\label{tab:rerun}

\begin{tabular}{@{}lccc@{}}
\toprule
Model & Item $\times$ input pairs & Identical codes & All nine codes identical\\
\midrule
Gemma 4 12B & 300 & 99.0\% & 93\%\\
Gemma 4 E4B & 300 & 99.8\% & 98\%\\
Qwen3-Omni & 300 & 97.3\% & 81\%\\
Qwen2.5-Omni & 300 & 99.1\% & 93\%\\
Nemotron-3 Nano Omni & 300 & 97.9\% & 85\%\\
MiniCPM-o 4.5 & 300 & 98.3\% & 88\%\\
\bottomrule
\end{tabular}

\end{appendixtable}

\begin{appendixtable}
\centering\footnotesize
\caption{Re-run: percentage of identical individual codes by route.}
\label{tab:rerunroutes}

\begin{tabular}{@{}lrrrrrr@{}}
\toprule
Model & Metadata & Transcript & Extr.\ text & Fr.\,+\,audio & Text\,+\,video & Describe\\
\midrule
Gemma 4 12B & 99.8 & 99.6 & 99.1 & 98.2 & 97.8 & 99.8\\
Gemma 4 E4B & 99.3 & 100.0 & 99.8 & 99.8 & 99.8 & 100.0\\
Qwen3-Omni & 95.6 & 97.8 & 98.7 & 97.8 & 98.4 & 95.6\\
Qwen2.5-Omni & 98.2 & 99.6 & 98.4 & 99.8 & 99.8 & 98.9\\
Nemotron-3 Nano Omni & 98.4 & 95.8 & 98.4 & 98.9 & 98.0 & 97.6\\
MiniCPM-o 4.5 & 95.3 & 98.0 & 98.7 & 98.7 & 98.9 & 100.0\\
\bottomrule
\end{tabular}

\end{appendixtable}

\FloatBarrier
\needspace{0.75\textheight}
\section{Cost}\label{app:cost}

\begin{appendixtable}
\centering\footnotesize
\caption{GPU time per news video and cost per 1,000 news videos (all nine constructs in one call; describe first includes the description call). We time each batch and divide its runtime by the batch size. Hourly prices for one A100 GPU: University of Zurich computing rate, CHF 0.081; Amazon Web Services on demand, US\$3.43 (p4de.24xlarge with eight A100 80GB GPUs at US\$27.45 per hour, US East) \citep{AWS2026ec2pricing}; Microsoft Azure pay-as-you-go, US\$3.673 (NC24ads A100 v4 with one A100 80GB GPU, East US) \citep{Microsoft2026azurepricing}. We exclude transcription and OCR costs because we run these steps once and reuse the results across models.}
\label{tab:cost}

\begin{tabular}{@{}llrrrr@{}}
\toprule
 & & & \multicolumn{3}{c}{Cost per 1,000 videos}\\
\cmidrule(l){4-6}
Model & Route & GPU s per video & UZH (CHF) & AWS (US\$) & Azure (US\$)\\
\midrule
Gemma 4 12B & Extracted text & 0.26 & 0.006 & 0.25 & 0.27\\
 & Frames + audio & 1.11 & 0.025 & 1.06 & 1.13\\
 & Text + video & 1.21 & 0.027 & 1.16 & 1.24\\
 & Describe first & 1.42 & 0.032 & 1.35 & 1.45\\
\addlinespace
Gemma 4 E4B & Extracted text & 0.09 & 0.002 & 0.09 & 0.09\\
 & Frames + audio & 0.87 & 0.020 & 0.83 & 0.89\\
 & Text + video & 0.91 & 0.020 & 0.86 & 0.92\\
 & Describe first & 0.97 & 0.022 & 0.93 & 0.99\\
\addlinespace
Qwen3-Omni & Extracted text & 0.13 & 0.003 & 0.13 & 0.14\\
 & Frames + audio & 0.36 & 0.008 & 0.34 & 0.36\\
 & Text + video & 0.41 & 0.009 & 0.39 & 0.41\\
 & Describe first & 0.75 & 0.017 & 0.72 & 0.77\\
\addlinespace
Qwen2.5-Omni & Extracted text & 0.14 & 0.003 & 0.13 & 0.14\\
 & Frames + audio & 0.60 & 0.013 & 0.57 & 0.61\\
 & Text + video & 0.70 & 0.016 & 0.67 & 0.71\\
 & Describe first & 0.80 & 0.018 & 0.77 & 0.82\\
\addlinespace
Nemotron-3 Nano Omni & Extracted text & 0.10 & 0.002 & 0.09 & 0.10\\
 & Frames + audio & 0.51 & 0.012 & 0.49 & 0.53\\
 & Text + video & 0.54 & 0.012 & 0.52 & 0.55\\
 & Describe first & 0.75 & 0.017 & 0.71 & 0.76\\
\addlinespace
MiniCPM-o 4.5 & Extracted text & 0.17 & 0.004 & 0.16 & 0.17\\
 & Frames + audio & 0.46 & 0.010 & 0.44 & 0.47\\
 & Text + video & 0.52 & 0.012 & 0.50 & 0.53\\
 & Describe first & 0.68 & 0.015 & 0.65 & 0.69\\
\bottomrule
\end{tabular}

\end{appendixtable}

We run all inference on single-A100 instances across three sessions. Including model downloads, loading, data transfer, and idle time, the complete study used about 32 hours of A100 time. At the University of Zurich computing rate this corresponds to about CHF 2.60; at the on-demand cloud prices above, to about US\$110 (Amazon Web Services) to US\$118 (Microsoft Azure).

\FloatBarrier
\needspace{0.30\textheight}
\section{Inferential analysis by coefficient}\label{app:infer}

\begin{appendixtable}
\centering\footnotesize
\caption{Engagement regression: human-code estimates and range of automated estimates across six models. OLS of log(1 + views) on all focal constructs with outlet fixed effects, HC1 standard errors, $n=599$. $^a$Automated estimates with the same sign as the human estimate, out of 18 (6 models $\times$ 3 routes).}
\label{tab:inferfull}
\fitwidth{
\begin{tabular}{@{}lccccc@{}}
\toprule
 & Human codes & \multicolumn{3}{c}{Automated estimates, range across six models} & \\ \cmidrule(lr){3-5}
Coefficient & $\hat\beta$ (SE) & Extracted text & Frames + audio & Native video & Sign agreement$^a$\\
\midrule
Negative news & .63 (0.29) & $-$.31 to .42 & .14 to .59 & .17 to .52 & 17/18\\
Conflict & .79 (0.30) & $-$.26 to .33 & $-$.07 to .27 & $-$.00 to .49 & 14/18\\
Scope & $-$.32 (0.33) & $-$.11 to .80 & $-$1.74 to .31 & $-$.57 to .30 & 8/18\\
Call to interact (TikTok) & $-$.13 (0.29) & $-$.37 to .31 & $-$.23 to .22 & $-$.08 to .22 & 7/18\\
Call to interact (off-platform) & $-$.37 (0.33) & $-$.56 to .00 & $-$.40 to .70 & $-$1.16 to .32 & 13/18\\
Other people's voice & $-$.19 (0.28) & $-$.56 to .32 & $-$.69 to $-$.00 & $-$.55 to .16 & 14/18\\
Geographic scope: domestic & $-$.50 (0.33) & $-$.76 to .19 & $-$.84 to .51 & $-$.55 to .30 & 14/18\\
Geographic scope: foreign & $-$.71 (0.36) & $-$1.02 to $-$.06 & $-$.73 to $-$.23 & $-$.61 to $-$.20 & 18/18\\
\bottomrule
\end{tabular}
}
\end{appendixtable}

\begin{appendixtable}
\centering\footnotesize
\caption{Accuracy and coefficient recovery within each coefficient. $\rho$: Spearman correlation across pipelines between the construct's macro-F1 and the absolute deviation of the automated from the human-code coefficient (negative: more accurate pipelines recover the estimate better). Descriptive: pipelines share items, models, and routes.}
\label{tab:inferestimand}
\fitwidth{
\begin{tabular}{@{}lrrrrrrr@{}}
\toprule
 & \multicolumn{2}{c}{Human codes} & \multicolumn{3}{c}{18 pipelines} & \multicolumn{2}{c}{Highest-F1 route}\\
\cmidrule(lr){2-3}\cmidrule(lr){4-6}\cmidrule(l){7-8}
Coefficient & $\hat\beta$ & SE & $\rho$ (F1, error) & Median error (SE units) & $\rho$, 36 pipelines & Closest (of 6) & Regret (SE units)\\
\midrule
Negative news & .63 & .29 & .44 & 1.22 & .34 & 1 & 0.78\\
Conflict & .79 & .30 & $-$.46 & 2.14 & $-$.09 & 5 & 0.00\\
Scope & $-$.32 & .33 & $-$.06 & 1.12 & $-$.01 & 1 & 0.74\\
Call to interact (TikTok) & $-$.13 & .29 & $-$.30 & 0.75 & $-$.31 & 1 & 0.28\\
Call to interact (off-platform) & $-$.37 & .33 & $-$.52 & 0.49 & $-$.64 & 1 & 0.57\\
Other people's voice & $-$.19 & .28 & $-$.21 & 0.88 & $-$.09 & 4 & 0.00\\
Geographic scope: domestic & $-$.50 & .33 & .41 & 1.05 & .44 & 4$^a$ & 0.00\\
Geographic scope: foreign & $-$.71 & .36 & .23 & 0.73 & .29 & 4$^a$ & 0.00\\
\bottomrule
\multicolumn{8}{@{}p{\textwidth}@{}}{\footnotesize Error: $|\hat\beta_{\text{auto}}-\hat\beta_{\text{human}}|$ divided by the human estimate's SE. 18 pipelines: six models $\times$ extracted text, frames + audio, native video; 36: all six routes used in the regression. Closest: number of models for which the route with the highest construct macro-F1 also has the smallest coefficient error among the three routes (chance: 2 of 6); regret: median excess error of that route over the smallest error. $^a$Geographic scope: one construct, error averaged over its two coefficients.}
\end{tabular}
}
\end{appendixtable}

\FloatBarrier
\section{Prompts}\label{app:prompts}
We use the same prompt template for every route within a corpus, changing only the one-sentence input description and the supplied material. Below, we provide the full news prompt for the extracted-text route; the input descriptions for all routes; the hate, target-group, and sarcasm prompts for frames + audio; and the construct-agnostic description prompt for describe first. We retain the internal route codes: C0 metadata only, Tasr transcript, Tgold human transcript, T extracted text, V frames + audio, Vnat16 native video, Vraw raw video, Vmute frames without audio, B text + video, and D describe first. Angle brackets mark item material.
\subsection*{News on TikTok, extracted-text route}
\begin{promptbox}
\begin{Verbatim}[fontsize=\promptfont]
You are a trained content analyst coding a German-language news video posted on TikTok by a
news outlet, following a published content-analysis codebook.
You receive the metadata, an automatic speech transcript of the audio track, and the on-screen
text extracted by optical character recognition (OCR). You do not see the video.

=== CODEBOOK ===
V6 Visual set-up. How is the video visually set up?
1 = Journalist-centric: A journalist/moderator is nearly constantly visible and in the
foreground.
2 = Live-footage with journalistic elements: One or multiple shot(s) of live-footage including
journalistic elements, such as a journalist or someone interviewing people.
3 = Live-footage without journalistic elements: One or multiple shot(s) of live-footage from an
event without a journalist visually present anywhere in the video.

V8b Other peoples' voice (overlap and/or in-situ). Is the voice of people other than the
journalist/moderator used in the video? 1 = used, 0 = not used. (Ignore the outlet endcard.)

V8d Music. Is music used in the video? 1 = used, 0 = not used. (Ignore the outlet endcard.)

V9b Call-to-interact on TikTok: visual- or audio-based prompts to the audience to interact with
the video or outlet on TikTok. This can include prompts to like, share, comment, discuss,
and/or share their thoughts on the topic in the video, e.g., "What do you think about this?" or
"Please share this video". Interactions with the outlet refer to subscribing to the channel,
following or clicking on their profile, that can also be purely visual or via an animation
(clicking on the subscribe button). 1 = present, 0 = not present.

V9c Call-to-interact outside of TikTok: visual- or audio-based prompts to the audience to
interact with the video or outlet that would result in the audience leaving TikTok. For example
by prompting the audience to go to the news outlet's website ("Klicke auf diesen Link, um mehr
zu erfahren") or to another platform ("Schau' mal auf unseren YouTube-Kanal nach"). 1 =
present, 0 = not present.

V12 Geographic scope. Compare the topic/location of the event of the video with the channel's
nationality (given in the material). Indications of the geographic scope of the video can be
explicit, i.e., mentioning a concrete location, or implicit, such as discussing the findings of
a representative study from a national research institute. Does the video deal with domestic or
foreign news?
0 = There is no concrete geographical scope to the topic, e.g., the video deals with a disease
and doesn't mention explicitly any specific location.
1 = The video deals with domestic news, e.g., srfnews (Swiss) shows footage from the rail
accident in the Gotthard Base Tunnel (in Switzerland) or zeitimbild (Austrian) is interviewing
Austrian politicians to be elected in the European Election.
2 = The video deals with foreign news, e.g., a video from zeitimbild (Austrian) that deals with
the process against Trump in the US or a video of tagesschau (German) dealing with an
earthquake in Mexico and interviewing German tourists in Mexico.

V15 Negative news. Stories with overall particularly negative topics such as death, injury,
defeat and loss (of a job, for example). They can be of economic, political, physical,
psychological or social nature. Guiding question: Does the story deal with bad news? Important:
code the topic, not the tonality. 1 = yes, 0 = no.

V16 Conflict. Stories concerning conflict such as controversies, arguments, splits, strikes,
fights, insurrections, and warfare. Is the focus of the story on the conflict?
0 = No, e.g., the video focusses on the fact that demonstrators vandalized buildings on their
march through a city center (what's the conflict?); report on the court decision on Sebastian
Kurz (not reported as conflict, but as decision); Italy will introduce smartphone-ban at
schools (not reported as conflict, but as decision).
1 = Yes, e.g., two opposing positions or opinions are mentioned about an event or a decision;
or a news outlet reports on Parisians demonstrating against the Olympic Games; reflection on
measurements taken during COVID in interview - including stating the opposing position.

V17 Scope. Stories that describe how many people are directly involved or potentially impacted
by the reported event. This includes both literal numbers mentioned with a verifiable source,
e.g., "thousands of people", or a single person, as well as figurative speeches of a group of
people, such as "all Netflix-viewers", "Every woman living in Germany" or "Germany". 1 = yes, 0
= no.

General rule: If the video summarizes multiple events, code the variables based on the clear,
first story in the video, excluding the introduction teaser.

=== MATERIAL ===
Outlet: srfnews (nationality of the channel: Switzerland)

Caption: <caption>

Hashtags: <hashtags>

Automatic speech transcript:
<automatic transcript>

On-screen text (OCR):
<on-screen text>

Code every variable strictly according to the codebook. Answer with one JSON object with the
keys "V6", "V8b", "V8d", "V9b", "V9c", "V12", "V15", "V16", "V17" and integer codes only.
\end{Verbatim}
\end{promptbox}
\needspace{.35\textheight}
\subsection*{Input descriptions (second sentence of every prompt)}
\begin{promptbox}
\begin{Verbatim}[fontsize=\promptfont]
C0: You receive only the post's metadata; no transcript and no video.
Tasr: You receive the metadata and an automatic speech transcript of the audio track. You do
not see the video.
Tgold: You receive the metadata and a human transcript of what is said. You do not see or hear
the video.
T: You receive the metadata, an automatic speech transcript of the audio track, and the on-
screen text extracted by optical character recognition (OCR). You do not see the video.
V: You receive the metadata, 16 frames sampled evenly from the video in temporal order, and the
video's audio track.
B: You receive the metadata, 16 frames sampled evenly from the video in temporal order, the
video's audio track, an automatic speech transcript, and on-screen text extracted by OCR.
D: You receive the metadata and a detailed description of the video written by an assistant who
watched and listened to it. You do not see the video yourself.
Vnat16, Vraw: You receive the metadata and the video with its audio track.
Vmute: You receive the metadata and 16 frames sampled evenly from the video in temporal order.
There is no audio.
\end{Verbatim}
\end{promptbox}
\needspace{.40\textheight}
\subsection*{HateMM, frames + audio route}
\begin{promptbox}
\begin{Verbatim}[fontsize=\promptfont]
You are a trained content analyst coding a video posted on the video platform BitChute,
following an annotation guideline for hateful content.
You receive the metadata, 16 frames sampled evenly from the video in temporal order, and the
video's audio track.

=== CODEBOOK ===
hate: Is this video hateful? A video is hateful if "it promotes discrimination or disparages or
humiliates an individual or group of people on the basis of the race, ethnicity, or ethnic
origin, nationality, religion, disability, age, veteran status, sexual orientation, gender
identity etc." 1 = hateful, 0 = not hateful.

=== MATERIAL ===
Metadata: none available.

[The 16 video frames and audio track are attached.]

Code every variable strictly according to the codebook. Answer with one JSON object with the
keys "hate" and integer codes only.
\end{Verbatim}
\end{promptbox}
\needspace{.40\textheight}
\subsection*{HateMM target groups, frames + audio route}
The two target variables are coded together in a separate call from hate.
\begin{promptbox}
\begin{Verbatim}[fontsize=\promptfont]
You are a trained content analyst coding a video posted on the video platform BitChute,
following an annotation guideline that records which communities a video targets.
You receive the metadata, 16 frames sampled evenly from the video in temporal order, and the
video's audio track.

=== CODEBOOK ===
target_black: Does the video target or concern Black people as a community (for example by
talking about, depicting, or attacking them)? 1 = yes, 0 = no.

target_jewish: Does the video target or concern Jewish people as a community (for example by
talking about, depicting, or attacking them)? 1 = yes, 0 = no.

=== MATERIAL ===
Metadata: none available.

[The 16 video frames and audio track are attached.]

Code every variable strictly according to the codebook. Answer with one JSON object with the
keys "target_black", "target_jewish" and integer codes only.
\end{Verbatim}
\end{promptbox}
\needspace{.47\textheight}
\subsection*{MUStARD, frames + audio route}
\begin{promptbox}
\begin{Verbatim}[fontsize=\promptfont]
You are a trained content analyst coding a short utterance from an English-language TV sitcom,
following an annotation guideline for sarcasm.
You receive the metadata, 16 frames sampled evenly from the video in temporal order, and the
video's audio track.

=== CODEBOOK ===
sarcasm: Is the TARGET UTTERANCE sarcastic? An utterance is sarcastic when the speaker's
intended meaning differs from, typically is the opposite of, its literal meaning, usually to
mock, tease, or express irritation or contempt. Sarcasm can be signalled by the words
themselves, by incongruity with the preceding context, or by delivery (tone of voice, stress,
exaggerated intonation, facial expression). 1 = sarcastic, 0 = not sarcastic.

=== MATERIAL ===
Show: FRIENDS. Speaker of the target utterance: <speaker>.

Preceding dialogue (context, human transcript):
<speaker>: <preceding line>
...

[The 16 video frames of the target utterance and audio track are attached.]

Code every variable strictly according to the codebook. Answer with one JSON object with the
keys "sarcasm" and integer codes only.
\end{Verbatim}
\end{promptbox}
\needspace{.40\textheight}
\subsection*{Description prompt (first stage of the describe-first route)}
The example below is for news. For HateMM (including the target-group calls), ``TikTok news video'' becomes ``video''; for MUStARD, it becomes ``short clip of one utterance from a TV sitcom''. The rest of the description instruction is unchanged.
\begin{promptbox}
\begin{Verbatim}[fontsize=\promptfont]
You see 16 frames sampled evenly from a TikTok news video (in temporal order) and its audio
track. Write a factual, exhaustive description for a researcher who cannot watch it. Use
exactly these headings:
SPEECH: transcribe what is said as completely as possible, in the original language, and say
who speaks (e.g., news presenter/journalist, interviewee, other person, voice-over).
ON-SCREEN TEXT: all text visible on screen, in the original language.
VISUALS: setting, people, and type of footage (e.g., presenter to camera, live footage,
graphics or maps, screenshots, stickers or emojis).
SOUND: music, sound effects, ambient noise, tone of voice.
Describe only; do not interpret or evaluate.
\end{Verbatim}
\end{promptbox}

\end{appendices}

\end{document}